\documentclass[a4paper,fleqn]{cas-dc}
\usepackage{algorithmic}
\usepackage{algorithm}
\usepackage{array}
\usepackage[authoryear]{natbib}
\usepackage{caption}
\usepackage{graphicx}
\usepackage{pifont}  
\usepackage{microtype}
\usepackage{caption}
\usepackage{booktabs}
\usepackage{multirow}
\usepackage{xcolor}
\usepackage[table]{xcolor}

\def\tsc#1{\csdef{#1}{\textsc{\lowercase{#1}}\xspace}}
\tsc{WGM}
\tsc{QE}
\tsc{EP}
\tsc{PMS}
\tsc{BEC}
\tsc{DE}

\begin{document}
\let\printorcid\relax	
\let\WriteBookmarks\relax
\def\floatpagepagefraction{1}
\def\textpagefraction{.001}
\shorttitle{ISPRS Journal of Photogrammetry and Remote Sensing}
\shortauthors{Yaoteng Zhang et~al.}

\title [mode = title]{STAR-IOD: Scale-decoupled Topology Alignment with Pseudo-label Refinement for Remote Sensing Incremental Object Detection}
\tnotetext[1]{This work is supported by the National Natural Science Foundation of China under Grant 62471394, 62306241, and 62576284.}
\author[1]{Yaoteng Zhang}
\ead{zhang_yt@mail.nwpu.edu.cn}

\author[2]{Qing Zhou}
\ead {mrazhou@mail.nwpu.edu.cn}
\author[2]{Junyu Gao}
\ead {gjy3035@gmail.com}

\author[2]{Qi Wang\corref{mycorrespondingauthor}}
\cortext[mycorrespondingauthor]{Corresponding author}
\ead{crabwq@gmail.com}

\address[1]{School of Computer Science, Northwestern Polytechnical University, Xi’an 710072, China}
\address[2]{School of Artificial Intelligence, OPtics and ElectroNics (iOPEN), Northwestern Polytechnical University, Xi’an 710072, China}

\begin{abstract}
Remote sensing imagery typically arrives in the form of continuous data streams. Traditional detectors often forget previously learned categories when learning new ones; therefore, research on Remote Sensing Incremental Object Detection (RS-IOD) is of great significance. However, existing methods largely overlook the intra-class scale variations prevalent in remote sensing scenes, which undermines the effectiveness of knowledge transfer and old knowledge preservation. Moreover, RS-IOD also suffers from missing annotations, which cause the model to misclassify old-class instances as background. To address these challenges, we propose a novel framework, STAR-IOD. First, we introduce a Subspace-decoupled Topology Distillation (STD) module to transfer structural knowledge, explicitly aligning inter-class topological relationships and mitigating intra-class representation discrepancies induced by scale shifts. Furthermore, we introduce the Clustering-driven Pseudo-label Generator (CPG), a plug-and-play module that leverages K-Means clustering to dynamically identify class-specific thresholds, thereby guaranteeing an accurate distinction between true positive targets and background noise and alleviating the issue of missing annotations for old classes. We also constructed two Remote Sensing Incremental Object Detection datasets, DIOR-IOD and DOTA-IOD  to facilitate research on RS-IOD. Extensive experiments demonstrate that our method outperforms state-of-the-art approaches by 1.7\% and 2.1\% mAP on DIOR-IOD and DOTA-IOD, respectively, effectively alleviating catastrophic forgetting while preserving strong detection performance on both base and novel classes. The code and dataset are released at: https://github.com/zyt95579/STAR-IOD.
\end{abstract}


\maketitle

\section{Introduction}
Deep learning has achieved remarkable progress in remote sensing object detection \citep{liu2025msdp,zou2025mosaic,lu2025legnet,yuan2025strip}, demonstrating strong potential in applications such as disaster monitoring, urban planning, and environmental protection. Conventional object detection paradigms \citep{han2021align,dai2022ao2,Marques_2025_CVPR,10530145,11093252} are typically predicated on the idealized assumption that training data follows a static distribution and is fully accessible in an offline setting. However, this premise stands in stark contrast to the open and dynamic nature of the real world, where data typically arrives sequentially in a continuous stream. When the model is updated on new tasks, conventional detection models tend to severely forget previously learned knowledge, a phenomenon known as catastrophic forgetting \citep{zhang2024learning,gda_iod,cermelli2022modeling}.
\begin{figure}
	\centering
	\includegraphics[width=1\columnwidth]{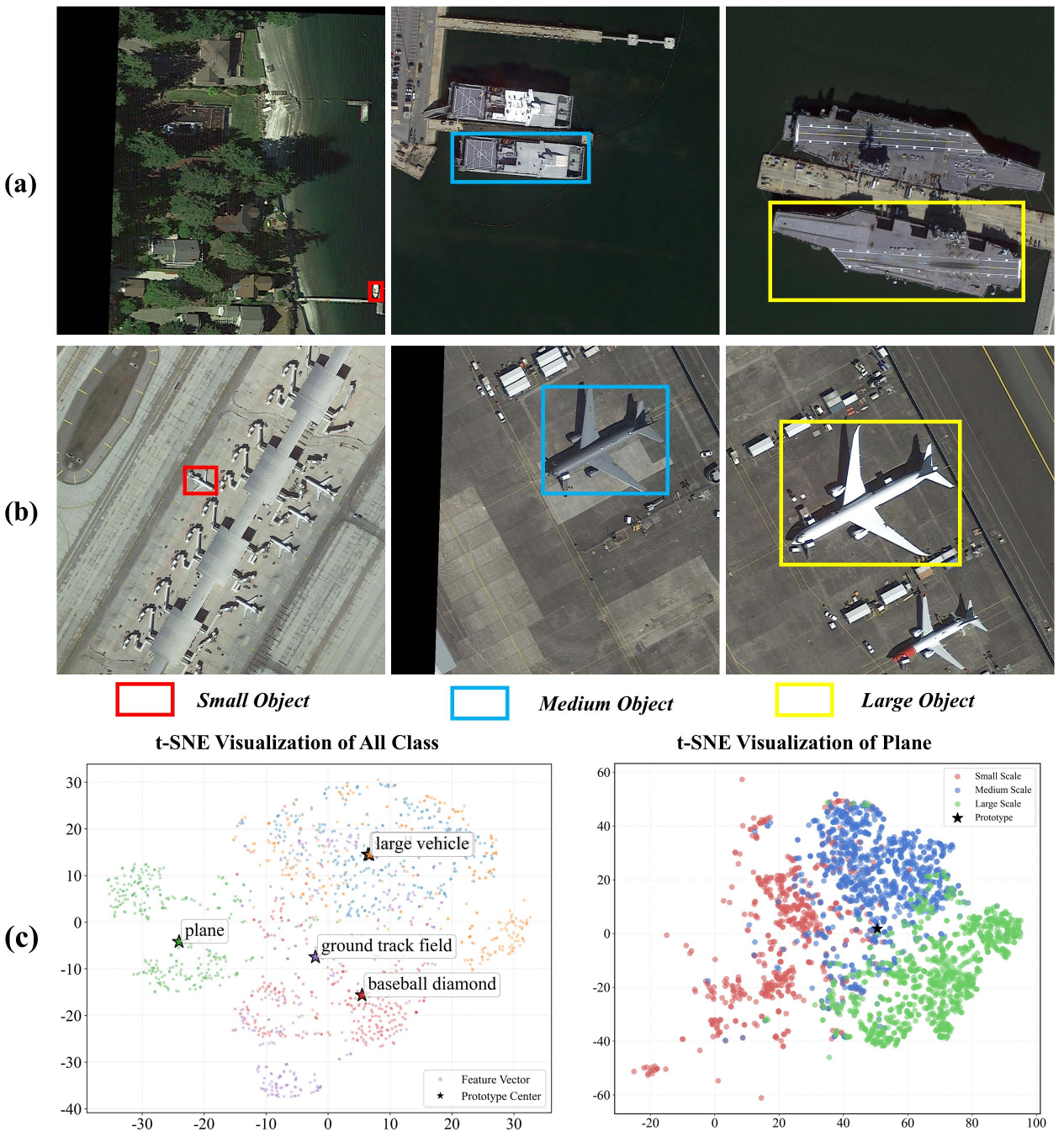}
	\caption{(a) Different scales of the ship in remote sensing images; (b) Different scales of the plane in remote sensing images; (c) t-SNE visualization of feature vectors in the feature space.}
	\label{Fig:1}
\end{figure}

To mitigate catastrophic forgetting, various Incremental Object Detection (IOD) methods~\citep{shmelkov2017incremental,li2017learning,chang2023detrdistill,kang2023alleviating} have been proposed, which can be broadly categorized into three paradigms: knowledge distillation, replay, and architecture adaptation. Knowledge distillation~\citep{wang2025gcd,liu2023continual} aims to preserve historical knowledge by transferring responses or feature representations from a frozen teacher model to the student. In contrast, replay-based approaches~\citep{liu2023augmented,rebuffi2017icarl,acharya2020rodeo,gao2023ddgr,yang2023one} maintain a subset of old samples or synthesized instances and alleviate forgetting via joint training. Architecture adaptation methods~\citep{wang2022dualprompt,wang2022learning,xu2025componential,smith2023coda,zhou2025learning} introduce task-specific modules or parameter isolation mechanisms to accommodate new knowledge while preserving previously learned representations. These approaches have achieved promising performance on IOD involving natural images. {However, unlike natural images, objects in remote sensing imagery typically exhibit highly challenging multi-scale characteristics. Specifically, remote sensing images are acquired from high-altitude platforms, where the spatial resolution varies with sensor altitude and imaging angles. Moreover, objects of the same category can differ significantly in their intrinsic physical sizes, as illustrated in Fig.~\ref{Fig:1}(a, b). Such intra-class scale variation makes it difficult for feature representations to form compact clusters in the feature space; instead, they exhibit pronounced scale dependency, as shown in Fig.~\ref{Fig:1}(c). This scale-dependent feature distribution reflects an inherent characteristic of remote sensing targets. Given this scale-dependent feature distribution, directly applying distillation or feature alignment strategies tends to indiscriminately compress scale-variant features into a unified representation. This indiscriminate feature compression not only overlooks the inherent intra-class scale diversity but also risks blurring inter-class boundaries and disrupting the inter-class topological structure. Consequently, it ultimately hinders the model’s ability to balance old knowledge retention and new knowledge acquisition in RS-IOD.}

Another critical challenge in RS-IOD stems from missing annotations \citep{kim2024sddgr,cermelli2022modeling,luo2025gradient,mo2024bridge}. In IOD settings, annotations are provided exclusively for newly introduced classes, causing instances of previously learned classes within new data to be treated as background. This foreground–background ambiguity introduces erroneous supervisory signals, suppressing old-class features and accelerating catastrophic forgetting. {Pseudo-labeling methods \citep{liu2023continual, bhatt2024preventing, wangpsedet} are commonly employed to recover missing annotations for old classes; however, their effectiveness in RS-IOD remains underexplored. In remote sensing scenarios, affected by complex backgrounds, dense object distributions, and frequent occlusions, pseudo-labels inevitably contain substantial noise, which significantly reduces the reliability of pseudo-supervision. These noisy pseudo-labels introduce erroneous supervisory signals that interfere with feature representation learning and are further amplified during the distillation process, thereby exacerbating the forgetting of previously learned classes.}
Furthermore, RS-IOD still suffers from the absence of standardized datasets and unified evaluation protocols, severely hindering fair comparisons among existing methods and restricting further investigation into this challenging problem.

To address the challenges of \textbf{intra-class scale variations} and \textbf{noisy pseudo-labels}, we propose a novel Remote Sensing Incremental Object Detection framework, named STAR-IOD. {STAR-IOD introduces a Subspace-Decoupling Topological Distillation (STD) module to enforce structured knowledge alignment between the teacher and student models. Specifically, STD employs Scale-adaptive Instance Partitioning (SIP) to divide object instances into small-, medium-, and large-scale subspaces according to their bounding box areas. Operating on the decoder output features, STD constructs inter-class topological matrices within each scale-specific subspace and aligns the student inter-class topology with that of the teacher model. This design effectively alleviates cross-scale feature interference caused by scale variations in remote sensing imagery, enabling the student model to preserve stable and discriminative old-class relational knowledge during incremental learning.}
{Second, to mitigate the adverse impact of noisy pseudo-labels in remote sensing scenarios, STAR-IOD integrates a Clustering-driven Pseudo-label Generator (CPG) to distinguish reliable pseudo-labels from noisy ones, thereby effectively reducing the bias introduced by noisy supervision during model optimization. Specifically, CPG reformulates confidence threshold estimation as a clustering problem and applies the K-Means algorithm to perform binary clustering in the confidence space, dynamically determining an adaptive decision boundary.} Furthermore, to bridge the gap in datasets and evaluation benchmarks for RS-IOD, we construct two dedicated datasets and design standardized incremental splitting strategies along with unified evaluation pipelines. These efforts establish a reproducible experimental foundation, which we hope will serve as a valuable benchmark for future research in this field. The main contributions of this work are summarized as follows:

(1) We introduce Subspace-decoupled Topology Distillation, which mitigates biased knowledge transfer induced by intra-class scale variations by aligning inter-class topological structures within explicitly decoupled scale-adaptive subspaces.

(2) We develop a Clustering-driven Pseudo-label Generator that estimates adaptive confidence thresholds via clustering, effectively mitigating noise interference from unreliable pseudo-labels.

(3) We construct two dedicated RS-IOD benchmarks, DIOR-IOD and DOTA-IOD, and design standardized incremental evaluation protocols to facilitate fair comparisons and reproducible future research.

(4) We propose STAR-IOD, a novel framework for RS-IOD, which consistently outperforms SOTA methods on both DIOR-IOD and DOTA-IOD, achieving improvements of 1.7\% $mAP$ and 2.1\% $mAP$, respectively.

\section{Related work}
\label{secrw}
\subsection{Object Detection}
In recent years, object detection has achieved remarkable progress. Two-stage detectors, such as R-CNN~\citep{girshick2015region}, Fast R-CNN~\citep{girshick2015fast}, and Faster R-CNN~\citep{ren2015faster}, adopt a region proposal paradigm followed by region-wise classification and bounding box regression, attaining high detection accuracy at the cost of substantial computational overhead.
To improve efficiency, one-stage detectors—including YOLO~\citep{14,33,32,31}, SSD~\citep{liu2016ssd}, and RetinaNet~\citep{lin2017focal}—directly predict object categories and bounding boxes from dense feature maps, enabling real-time inference while maintaining competitive performance.
In recent years, Transformer-based detectors have reshaped object detection by introducing a new paradigm. DETR~\citep{carion2020end} reformulates detection as a set prediction problem, eliminating hand-crafted components such as anchor boxes and non-maximum suppression (NMS). Building upon this framework, subsequent methods—including Deformable DETR~\citep{zhu2020deformable}, DN-DETR~\citep{li2022dn}, and DINO~\citep{zhang2022dino}—further enhance convergence speed and detection accuracy through deformable attention mechanisms, denoising training strategies, and improved matching schemes.
Owing to their unified formulation and architectural flexibility, these detectors have been widely adopted in both natural image and remote sensing applications. Nevertheless, most existing object detection models are trained under a closed-world assumption, where all training data are available in advance and the data distribution remains fixed. Therefore, when continuously optimized on new data, traditional object detectors struggle to maintain detection accuracy on previously learned categories, which has motivated research in IOD methods.

\subsection{Incremental Object Detection}
Owing to the absence of samples and annotations for previously learned classes, object detectors are prone to catastrophic forgetting when trained on new tasks. Early incremental object detection approaches, such as ILOD~\citep{shmelkov2017incremental}, adopt the Learning without Forgetting (LwF)~\citep{li2017learning} paradigm to alleviate forgetting through knowledge distillation.
Subsequent methods have explored complementary strategies to enhance knowledge retention. ABR~\citep{liu2023augmented} combines old-class bounding box replay with RoI-level distillation, achieving improved preservation of historical object representations. In contrast to replay-based approaches, ERD~\citep{feng2022overcoming} mitigates forgetting by independently distilling classification and regression responses. MMA~\citep{cermelli2022modeling} further proposes an Unbiased Knowledge Distillation strategy that explicitly models missing annotations, balancing old knowledge retention with new knowledge acquisition.
Another line of research focuses on pseudo-label generation to recover missing supervision for old classes. OW-DETR~\citep{gupta2022ow} introduces an attention-based pseudo-labeling mechanism to identify unannotated old-class instances, while CL-DETR~\citep{liu2023continual} and MD-DETR~\citep{bhatt2024preventing} leverage threshold-based pseudo-label supervision to alleviate forgetting. However, due to substantial variations in confidence distributions across categories, fixed confidence thresholds often fail to generalize effectively across classes, limiting the reliability of pseudo-labels.
Compared to natural image scenarios, research on RS-IOD remains relatively limited, largely due to complex imaging conditions and diverse background patterns. Inspired by the LwF paradigm, FPN-IL~\citep{chen2020incremental} introduces an extended branch architecture that alleviates catastrophic forgetting by distilling classification and regression knowledge from a frozen teacher model. Nevertheless, this method overlooks the label ambiguity inherent in incremental remote sensing settings, where instances of newly introduced classes may appear as background in historical data. To mitigate this issue, SDCID~\citep{ruan2022class} proposes a selective distillation strategy that employs masking mechanisms to prevent the teacher network from producing misleading background supervision for novel targets.
Despite these efforts, existing studies have neither established standardized dataset partitioning protocols for RS-IOD nor developed dedicated feature modeling strategies to explicitly address severe multi-scale variations and complex background interference. These limitations hinder further performance improvements and motivate the development of more robust RS-IOD frameworks.

\section{Preliminaries}
\subsection{Incremental Object Detection} 
The training process of IOD is organized into $n$ sequential stages, where each stage introduces a disjoint set of novel categories. Let the overall class set be $\mathcal{C} = \{\mathcal{C}_1, \mathcal{C}_2, \ldots, \mathcal{C}_n\}$ with $C_i \cap C_j = \emptyset, \forall i \neq j$. At stage $t$, the detector is trained on the dataset $\mathcal{D}_t = \{X_t, Y_t\}$, where $X_t$ represents the input images and $Y_t$ the corresponding annotations of classes in $C_t$. Although images may contain objects from any category in $\mathcal{C}$, only instances of $C_t$ are annotated. The objective is to adapt the detector from $\mathcal{M}_{t-1}$ to $\mathcal{M}_t$ using only the current dataset $\mathcal{D}_t$, without accessing previous datasets $\{\mathcal{D}_1, \ldots, \mathcal{D}_{t-1}\}$, while preventing catastrophic forgetting and maintaining strong performance on previously learned classes $\{C_1, \ldots, C_{t-1}\}$.

\subsection{Grounding DINO}

Grounding DINO~\citep{liu2023grounding} is a vision-language detector designed for phrase grounding, aiming to associate visual objects with textual phrases. It consists of a visual backbone $f_v$ and a language backbone $f_l$, which are responsible for extracting modality-specific representations. Given an image-text pair $(I, T)$, the feature enhancer $f_e$ performs cross-modal fusion to associate visual and textual features:
\begin{equation}
    V, W = f_e(f_v(I), f_l(T))
\end{equation}

After fusion, the aligned image and text embeddings are denoted as $V, W$ respectively. The cross-modal alignment score is computed using cosine similarity:
\begin{equation}
    S(V_i, W_j) = \frac{v_i \cdot w_j^\top}{\|v_i\|\|w_j\|}
\end{equation}

The query selection module initializes reference points from the alignment scores $S(V, W)$, which serve as the positional components of learnable object queries. A cross-modal decoder further refines these queries, producing a set of query embeddings $Q = \{q_i\}_{i=1}^{N} \in \mathbb{R}^{N \times D}$, where $N$ is the number of queries and $D$ is the query embedding dimension. The decoder output $\hat{y} = \{\hat{y}_i\}_{i=1}^N$ is a sequence of object predictions $\hat{y}_i = (\hat{s}_i, \hat{b}_i)$ including logits and bounding boxes, where $\hat{s}_i = S(q_i, W)$ and $\hat{b}_i$ is predicted by the regression head.

During training, each ground-truth instance is assigned to a query by minimizing the matching cost between predictions and annotations. The optimal assignment is the solution to:
\begin{equation}
    \hat{\sigma} = \operatorname*{arg\,min}_{\sigma} \sum_{i=1}^{N} \mathcal{L}_{{match}}(\hat{y}_{\sigma(i)}, y_i)
\end{equation}
where $\sigma$ denotes a permutation over $N$ predictions and $\hat{\sigma}$ represents the optimal matching. $y_i = (s_i, b_i)$ is the $i$-th GT. $\mathcal{L}_{{match}}$ is a pair-wise matching cost defined as:
\begin{equation}
    \mathcal{L}_{{match}}(\hat{y}_{\sigma(i)}, y_i) = \mathcal{L}_{{align}}(\hat{s}_{\sigma(i)}, s_i) + \mathcal{L}_{{reg}}(\hat{b}_{\sigma(i)}, b_i)
\end{equation}

Since the number of ground-truth objects in an image is typically smaller than $N$. Predictions that do not match any ground-truth instance are considered negative samples, whose corresponding alignment targets are padded to zero, denoted as $\phi$. Accordingly, the overall DETR-style training loss is defined as:
\begin{equation}
    \mathcal{L}_{{detr}}(\hat{y}, y) = \sum_{i=1}^{N} \left[ \mathcal{L}_{{align}}(\hat{s}_{\sigma(i)}, s_i) + \mathbb{1}_{\{s_i \neq \phi\}} \mathcal{L}_{{reg}}(\hat{b}_{\hat{\sigma}(i)}, b_i) \right]
\end{equation}

\begin{figure*}
    \centering
    \includegraphics[width=0.8\linewidth]{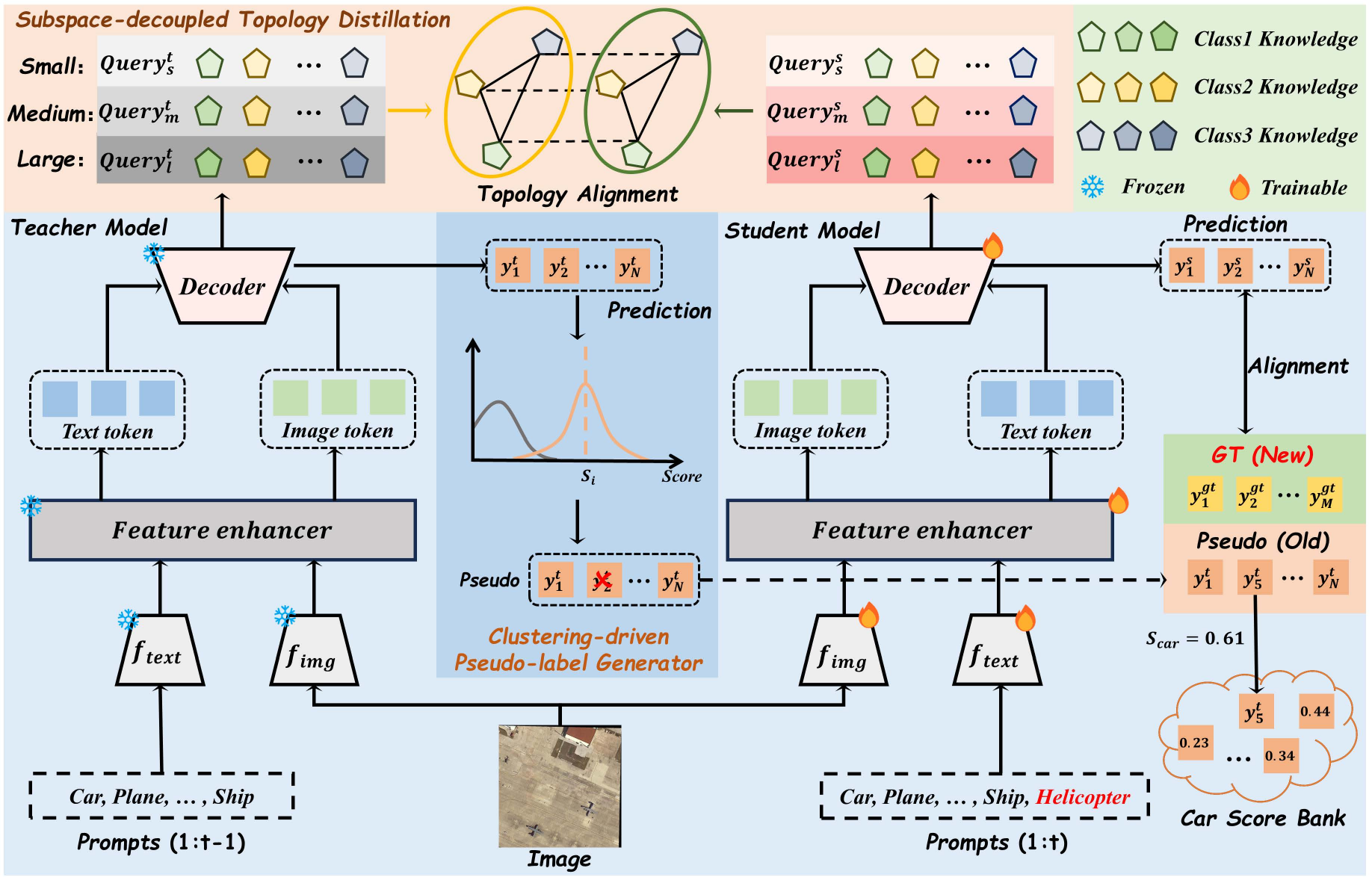}
    \caption{Overview of the proposed {STAR-IOD}. STAR-IOD preserves historical knowledge by aligning the decoder outputs of the student and teacher models via {Subspace-decoupled Topology Distillation}. In parallel, it incorporates a {Clustering-driven Pseudo-label Generator} to adaptively generate pseudo-labels for previously learned classes, thereby providing consistent and reliable supervision throughout the incremental learning process.}
    \label{overall}
\end{figure*}

\section{Methodology}\label{secmethod}
\subsection{Overall Architecture of STAR-IOD}
Architecturally grounded in Grounding DINO, the STAR-IOD framework combats catastrophic forgetting in RS-IOD via the synergy of two pivotal modules: Subspace-decoupled Topology Distillation (STD) and a Clustering-driven Pseudo-label Generator (CPG), as depicted in Fig.~\ref{overall}. Specifically, STD enforces strict feature-space alignment between teacher and student decoders, whereas CPG rectifies supervision signals to guarantee foreground consistency and reliability throughout the incremental learning process. During the incremental phase, STD exploits scale priors from the teacher's predictions to stratify query features into decoupled subspaces corresponding to small, medium, and large object scales. Within each subspace, it reconstructs and aligns inter-class topological relationships, thereby facilitating the precise transfer of fine-grained structural knowledge across varying scales. To counteract the prevalence of missing annotations and label noise in remote sensing imagery, CPG discards rigid fixed-threshold heuristics in favor of an adaptive, data-driven mechanism. By maintaining a class-specific confidence memory bank and performing intra-class clustering, CPG derives adaptive filtering thresholds for each category, enabling reliable supervision even for hard-to-detect objects. This robust design empowers the model to adapt seamlessly to the complex backgrounds and extreme scale variations characteristic of remote sensing imagery.

\subsection{Subspace-decoupled Topology Distillation}
\label{sec:strd}
\begin{figure*}
    \centering
    \includegraphics[width=0.7\linewidth]{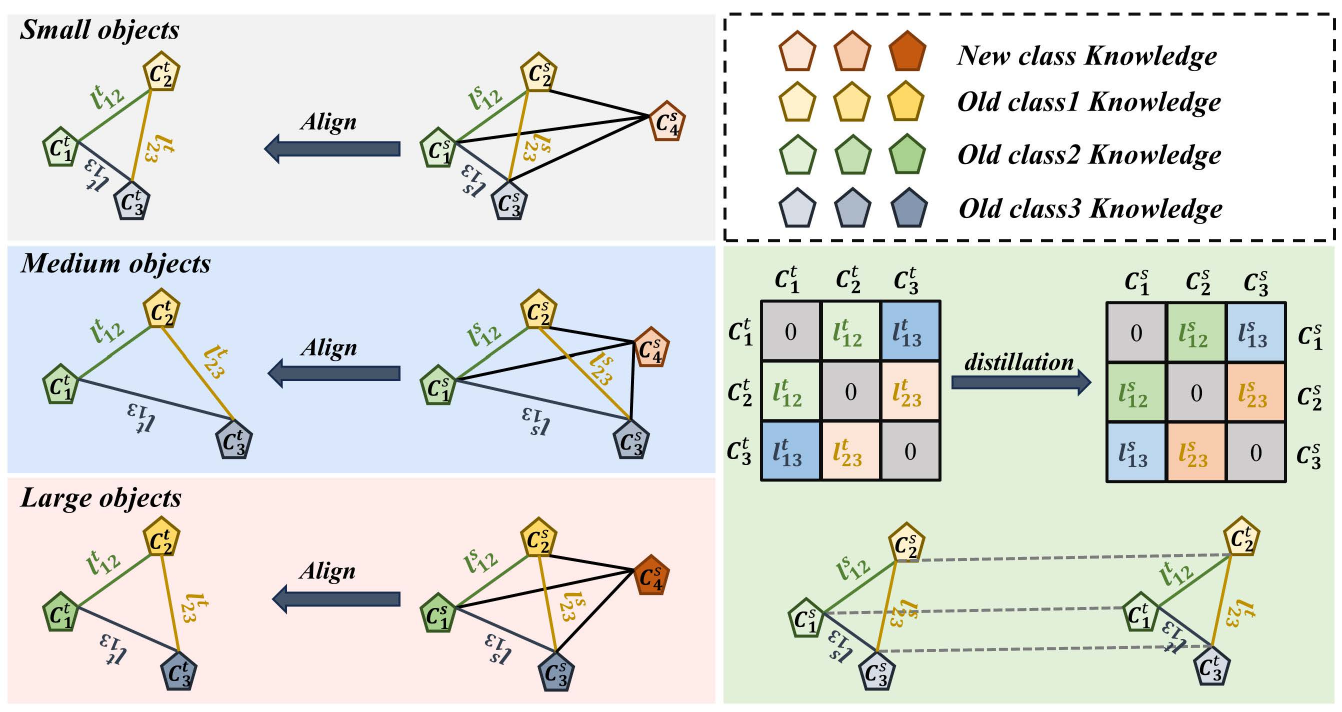}
    \caption{Overview of the proposed STD. STD enforces the alignment of inter-class topological relationships within three decoupled, scale-specific subspaces. It independently constrains inter-class distances between the teacher and student models across small, medium, and large scales, thereby preserving the geometric consistency of structured knowledge during incremental learning.}
    \label{ITD}
\end{figure*}
Remote sensing object features exhibit pronounced scale dependency. However, existing IOD methods typically model intra-class features using a single class prototype, thereby overlooking the representation discrepancies induced by intra-class scale variations. To overcome these limitations, we propose a 
Subspace-decoupled Topology Distillation. As depicted in Fig.~\ref{ITD}, STD stratifies features into distinct scale subspaces (small, medium, large) to reconstruct and align class-wise topological structures between teacher and student models. This process consists of three key phases: (1) Scale-adaptive Instance Partitioning, (2) Context-injected Prototype Aggregation, and (3) Topological Relation Alignment.

\noindent\textbf{Scale-adaptive Instance Partitioning:} {Due to intra-class scale variations, feature representations exhibit a distinct scale-dependent distribution, making it difficult for features of the same class to form compact clusters. This consequently weakens the discriminative boundaries between categories and hinders the learning of discriminative feature structures. Therefore, we decouple features of different scales into mutually independent subspaces, thereby preserving the accuracy and stability of old-class knowledge distillation.} Let $\mathcal{B} = \{b_i\}_{i=1}^{N}$ denote the set of predicted bounding boxes and $\mathcal{F} = \{f_i\}_{i=1}^{N}$ be the corresponding instance query features, where $f_i \in \mathbb{R}^D$.

We begin by calculating the area $A(b_i)$ of each predicted bounding box. Based on predefined scale thresholds $\tau_s$ and $\tau_m$, the instance query features are stratified into three disjoint subsets:
\begin{equation} 
\mathcal{Q}_{k} = \{f_i \mid {cond}_k(A(b_i))\},\quad k \in \mathcal{S} = \{{s_s, s_m, s_l}\}
\end{equation}
where the selection conditions are defined as: ${cond}_{{s_s}}$ for $A(b_i) < \tau_s$, ${cond}_{{s_m}}$ for $\tau_s \leq A(b_i) < \tau_m$, and ${cond}_{{s_l}}$ for $A(b_i) \geq \tau_m$. In our implementation, we set $\tau_s = 1024$ and $\tau_m = 9216$. This multi-scale partitioning strategy ensures that the student mimics the teacher's relational reasoning within respective scale scopes, effectively mitigating the ambiguity arising from the extreme scale variance in remote sensing imagery.

\noindent\textbf{Context-injected Prototype Aggregation:}
{To construct the nodes of the topological graph, we construct prototypes from the pseudo-labeled old classes present in the current training batch. Instead of simple arithmetic averaging, we employ a confidence-weighted aggregation strategy. This approach effectively attenuates the influence of low-quality predictions and background noise, ensuring that the resulting old-class prototypes better capture discriminative object features.} 

{For a specific scale $k$ and each category present in the current batch, the class prototype $\mathbf{p}_{c}^{k}$ is computed as:}

\begin{equation}
    \mathbf{p}_{c}^{k} = \sum_{i \in \mathcal{Q}_k} \mathbb{1}(y_i = c) \cdot \frac{s_i}{\sum_{j} s_j} \cdot f_i,
    \quad c \in \mathcal{C}_{1:(t-1)}
\end{equation}
where $\mathbb{1}(\cdot)$ is the indicator function, $y_i$ is the pseudo-label, and $s_i$ is the classification confidence score.

\noindent\textbf{Geographical Context Injection:} 
In remote sensing imagery, the semantic coupling between objects and their environmental context provides a critical inductive bias. To exploit this, we introduce a background anchor node to capture global-scale context. Let $F_{{img}} \in \mathbb{R}^{H \times W \times D}$ be the global image feature map. The background prototype $\mathbf{p}_{bg}$ is derived via global average pooling:
\begin{equation}
    \mathbf{p}_{{bg}} = \frac{1}{H \times W} \sum_{x=1}^{H} \sum_{y=1}^{W} F_{{img}}(x,y)
\end{equation}

This prototype is integrated into the graph as a static reference anchor, facilitating the explicit modeling of object-background dependencies within the topological structure.

\noindent\textbf{Topology Construction and Alignment:}
{The inter-class topological structure at each scale $k$ is encapsulated by a relation matrix $\mathbf{M}^k$,  where each entry $\mathbf{M}_{u,v}^k = \lVert \mathbf{p}_u^k - \mathbf{p}_v^k \rVert_2$ quantifies the semantic discrepancy between class prototypes in the feature space, with $u, v \in \mathcal{C}_{1:(t-1)}$ denoting old classes. Both the class prototypes and the relation matrix are dynamically constructed from the subset of old classes present in the current training batch. Specifically, for each scale subspace $k$, prototypes are computed only for pseudo-labeled queries, while absent classes are excluded from both prototype construction and relation matrix computation. If the number of valid old classes in scale subspace $k$ is $N_k$, the relation matrix is constructed as $\mathbf{M}^k \in \mathbb{R}^{N_k \times N_k}$.}

To rigorously enforce topological consistency, we move beyond rigid geometric alignment by modeling inter-class relationships as a probabilistic distribution. We map the Euclidean distances into a semantic affinity distribution $\mathcal{P}^k$. Specifically, the probability of semantic correlation between category $u$ and $v$ is formulated using a Boltzmann-like distribution \citep{hopfield1985neural}:
\begin{equation}
    \mathcal{P}_{u,v}^k = \frac{\exp(-M_{u,v}^k / \tau)}{\sum_{j=1}^{C'} \exp(-M_{u,j}^k / \tau)}
\end{equation}
where $\tau$ is a temperature hyperparameter that modulates the sharpness of the distribution, thereby governing the model's sensitivity to subtle inter-class dependencies. This transformation ensures that smaller geometric distances in the embedding space translate into higher semantic affinities.

Consequently, the STD objective is formulated as the Kullback-Leibler (KL) divergence between the teacher's and student's affinity distributions. To balance the gradient magnitude at different temperatures, the loss is scaled by $\tau^2$:
\begin{equation}
    \mathcal{L}_{{STD}} = \sum_{k \in \mathcal{S}} \tau^2 \sum_{u=1}^{C'} \mathcal{D}_{{KL}}\left( \mathcal{P}_{T, u}^k \big\| \mathcal{P}_{S, u}^k \right)
\end{equation}
where $\mathcal{P}_{T, u}^k$ and $\mathcal{P}_{S, u}^k$ denote the affinity distributions of the $u$-th prototype for the teacher and student, respectively. By minimizing $\mathcal{L}_{\text{STD}}$, the student network is compelled to reconstruct the rigorous semantic topology and relational reasoning logic established by the teacher.

\subsection{Clustering-driven Pseudo-label Generator}
\label{CPG}
\begin{figure*}
    \centering
    \includegraphics[width=0.7\linewidth]{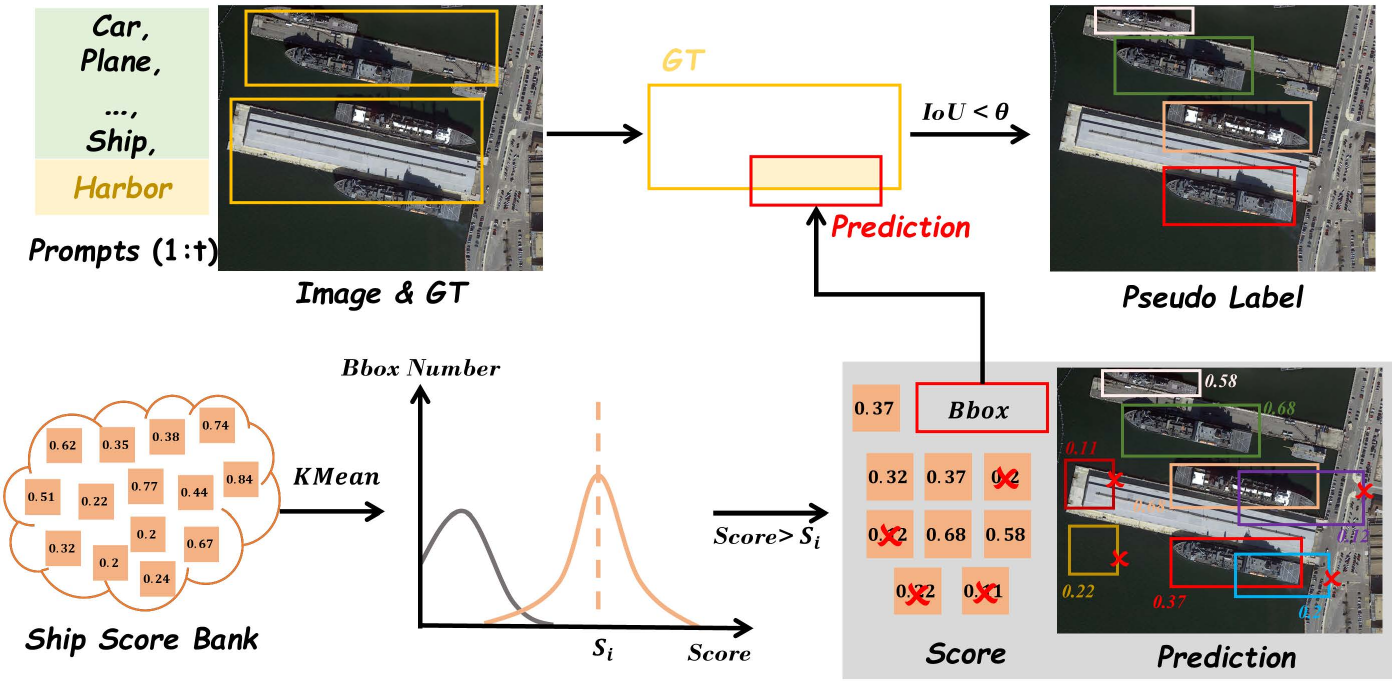}
    \caption{Overview of the proposed CPG. CPG leverages clustering to estimate adaptive thresholds for each category from prediction scores, thereby enabling the generation of high-quality pseudo-labels for previously learned classes.}
    \label{CPG1}
\end{figure*}
{In the IOD setting, the absence of annotations for previously learned classes inevitably biases the model to treat old-class instances as background, thereby precipitating catastrophic forgetting. This issue is further amplified in remote sensing scenarios, where complex backgrounds, dense object distributions, partial occlusion, large scale variations, and imaging noise lead to highly unstable confidence distributions. Conventional IOD approaches \citep{liu2023continual,bhatt2024preventing} typically generate pseudo-labels using a static confidence threshold, e.g., $\tau=0.4$. However, such a fixed-threshold strategy is not well suited to RS-IOD, since the confidence distributions of different categories and incremental stages may vary substantially, making it difficult to distinguish reliable old-class predictions from noisy pseudo-labels.}

{To overcome this limitation, we propose a Clustering-driven Pseudo-label Generator module, as shown in Fig.~\ref{CPG1}. Eschewing manual hyperparameter tuning, we reformulate the threshold determination as a clustering task within the 1D confidence space. This mechanism dynamically derives decision boundaries from the statistical distribution of model predictions. It is worth noting that CPG estimates confidence thresholds based on the outputs of a Transformer-based teacher model. Benefiting from its powerful self-attention mechanism, the detector has already captured and integrated rich global contextual information before producing the final prediction scores.}

\noindent\textbf{Dynamic Score Maintenance:} Given the sparsity of old-class instances within a single mini-batch, the statistical stability required for reliable clustering is often compromised. Therefore, we maintain a class-specific score bank $\mathcal{Q}_c$ for each category $c$.

Let $S_{c} = \{s_i \mid y_i=c, s_i > \delta_{min} \}$ denote the set of incoming prediction scores for class $c$ in the current batch that exceed a minimal candidate filter $\delta_{min}$. We update the bank as follows:
\begin{equation}
    \mathcal{Q}_c = {Concat}(\mathcal{Q}_c, S_{c})
\end{equation}
The bank length is bounded by $L_{cpg}$ to ensure the statistics reflect the model's current state while providing enough samples for clustering. In our experiments, $\delta_{min}$ is set to $0.3$, and $L_{cpg}$ is set to $20,000$.

\noindent\textbf{Clustering-guided Threshold Estimation:} Interestingly, low-confidence False Positives (FPs) and high-confidence True Positives (TPs) exhibit a bimodal distribution with varying inter-peak margins across classes. This phenomenon underscores the feasibility of using clustering to dynamically determine class-specific thresholds. To dynamically obtain the threshold, we employ a 1D K-Means clustering algorithm on the scalar values stored in $\mathcal{Q}_c$. Formally, we aim to partition the set $\mathcal{Q}_c$ into two clusters $\mathcal{C}_{l}$ and $\mathcal{C}_{h}$ by minimizing the within-cluster sum of squares:
\begin{equation}
    \min_{\mu_{l}, \mu_{h}} \sum_{s \in \mathcal{C}_{l}} (s - \mu_{l})^2 + \sum_{s \in \mathcal{C}_{h}} (s - \mu_{h})^2
\end{equation}
where $\mu_{l}$ and $\mu_{h}$ are the centroids of the low and high-confidence clusters, respectively. 

The adaptive threshold $\tau_c$ for category $c$ is defined as the lower bound of the high-confidence cluster $\mathcal{C}_{h}$:
\begin{equation}
    \tau_c = \min \{ s \mid s \in \mathcal{C}_{h} \}
\end{equation}

With the adaptive threshold $\tau_c$, we generate pseudo-labels from the teacher model's outputs. Let $\mathcal{O} = \{(b_i, s_i, c_i)\}$ be the set of raw predictions, where $b_i$, $s_i$, and $c_i$ denote the bounding box, confidence score, and category label, respectively. The initial set of pseudo-labels $\mathcal{P}_{r}$ is obtained by:
\begin{equation}
    \mathcal{P}_{r} = \{ (b_i, c_i) \mid s_i \geq \tau_{c_i} \}
\end{equation}

\noindent\textbf{Box De-duplication:} In incremental settings, the teacher model may detect objects that are already annotated in the ground truth set $\mathcal{G}$ of the current batch. To prevent duplicate supervision and ambiguous loss calculation, we perform a spatial filtering step. We compute the Intersection over Union (IoU) between every pseudo-box $b_p \in \mathcal{P}_{r}$ and the ground truth boxes $b_{gt} \in \mathcal{G}$. A pseudo-label is retained only if it does not significantly overlap with known annotations:
\begin{equation}
    \mathcal{P} = \{ (b_p, c_p) \in \mathcal{P}_{r} \mid \max_{b_{gt} \in \mathcal{G}} \text{IoU}(b_p, b_{gt}) < \theta_{nms} \}
\end{equation}
where $\theta_{nms}=0.7$ is the suppression threshold. The filtered pseudo-labels are then integrated into the training targets, enabling the student model to learn from both manually annotated instances and reliable pseudo-labels.

\subsection{Loss Function}\label{lf}
\noindent\textbf{Correspondence Response Distillation:} To ensure consistency, we employ the Correspondence Response Distillation (CRD) \citep{wang2025gcd} to transfer response-level knowledge from the teacher to the student. For classification alignment, CRD uses KL divergence to distill the teacher's prediction distribution. Specifically, the teacher's logits are mapped into probabilities via $P_i^{\mathit{old}} = \mathit{SoftMax}(\hat{s}_i^{\mathit{old}} / \tau)$, where $\tau$ serves as a temperature coefficient to smooth the distribution. The student's probabilities $P_i$ are derived identically.

The alignment distillation loss is formally defined as:
\begin{equation}
    \mathcal{L}_{{CRD}}^{{align}} = \sum_{i=1}^{N} \alpha_i \mathcal{D}_{{KL}}(P_i^{\mathit{old}} \| P_i),
\end{equation}
where $\alpha_i = \max_{c \in \mathcal{C}_{1:t-1}} \hat{s}_i^{\mathit{old}}(c)$ denotes the teacher's prediction probability over previously learned categories. This confidence-aware weighting mechanism suppresses background-dominated responses while emphasizing foreground queries that contain rich semantic information.

In addition, CRD incorporates regression distillation to preserve localization knowledge by enforcing geometric consistency between the teacher and student bounding box predictions. The regression loss follows the same formulation as the bounding box regression loss in the detection objective and is weighted by the same confidence factor $\alpha_i$:
\begin{equation}
    \mathcal{L}_{{CRD}}^{{reg}} = \sum_{i=1}^{N} \alpha_i \mathcal{L}_{{bbox}}(b_i^{\mathit{old}}, b_i).
\end{equation}

The total CRD loss is derived by aggregating the alignment and regression components across all decoder layers, formulated as:
\begin{equation}
    \mathcal{L}_{{CRD}} = \sum_{l=1}^{L} \left( \mathcal{L}_{{CRD}}^{{align}, l} + \mathcal{L}_{{CRD}}^{{reg}, l} \right),
\end{equation}
{where $L$ denotes the number of decoder layers.}

\noindent\textbf{Overall Loss:} To strike a balance between adaptability to new tasks and the preservation of old knowledge during the incremental learning process, the overall optimization objective of STAR-IOD is composed of three components: the fundamental detection loss, the subspace-decoupled topology distillation loss, and the correspondence response distillation loss. The total loss $\mathcal{L}$ is defined as follows:
\begin{equation}
    \mathcal{L}_{} = \mathcal{L}_{detr} + \lambda_{1}\mathcal{L}_{STD} + \mathcal{L}_{CRD}
    \label{eq:total_loss}
\end{equation}
In the experiments, $\lambda_{1}$ is set to 3.

\section{Experiment}\label{secexp}
In this section, we conduct extensive experimental comparisons between STAR-IOD and existing SOTA methods. It is worth noting that, due to the lack of publicly available RS-IOD methods, the selected comparison methods are general-purpose approaches. We evaluate the performance of STAR-IOD through both objective quantitative analyses and subjective visualization analyses, and further conduct ablation studies to verify the effectiveness of the proposed modules in the STAR-IOD.

\subsection{Implementation Details}
To ensure a fair comparison, we evaluate the proposed STAR-IOD method and all competing approaches on the same experimental platform. The platform consists of a server equipped with an NVIDIA L20 GPU. In our experiments, we adopt the AdamW optimizer with an initial learning rate of $1 \times 10^{-4}$ and a weight decay of $1 \times 10^{-4}$. All models are trained for 25 epochs per task with a batch size of 4.

\subsection{Datasets and Evaluation Metrics}

To address the lack of public datasets for RS-IOD, we construct and release two benchmarks based on the existing DIOR \citep{li2020object} and DOTA \citep{xia2018dota} datasets: DIOR-IOD and DOTA-IOD. The specific settings are as follows:

\noindent\textbf{DIOR-IOD} follows a two-step incremental setting, denoted as $A+B$, as shown in Table~\ref{tab:dior}. $A$ represents the number of classes introduced in phase $T_1$, and $B$ represents the number of classes introduced in phase $T_2$. Each phase contains 10 classes, and there is no class overlap between the two phases. {Beyond this standard two-step protocol, we further construct a more challenging multi-step setting on DIOR-IOD, denoted as $5+5+5+5$, to evaluate the robustness of our method under longer incremental sequences. The task sequence is defined as:}
{{Task 1 (Base)}: airplane, airport, bridge, service-area, toll-station;}
{{Task 2 (Inc 1)}: baseball field, basketball court, golf field, chimney, dam;}
{{Task 3 (Inc 2)}: ground track field, stadium, storage tank, tennis court, windmill;}
{{Task 4 (Inc 3)}: harbor, overpass, ship, train station, vehicle.}

\noindent\textbf{DOTA-IOD} (see Table~\ref{tab:dota}) utilizes a multi-step setting, dividing the dataset into 3 sequential tasks, each containing 5 classes. Given the significant imbalance in the number of instances across classes (long-tailed distribution), DOTA-IOD serves as a robust benchmark to evaluate the model's incremental learning performance and resistance to catastrophic forgetting in complex scenarios.

\begin{figure}
    \centering
    \includegraphics[width=1\linewidth]{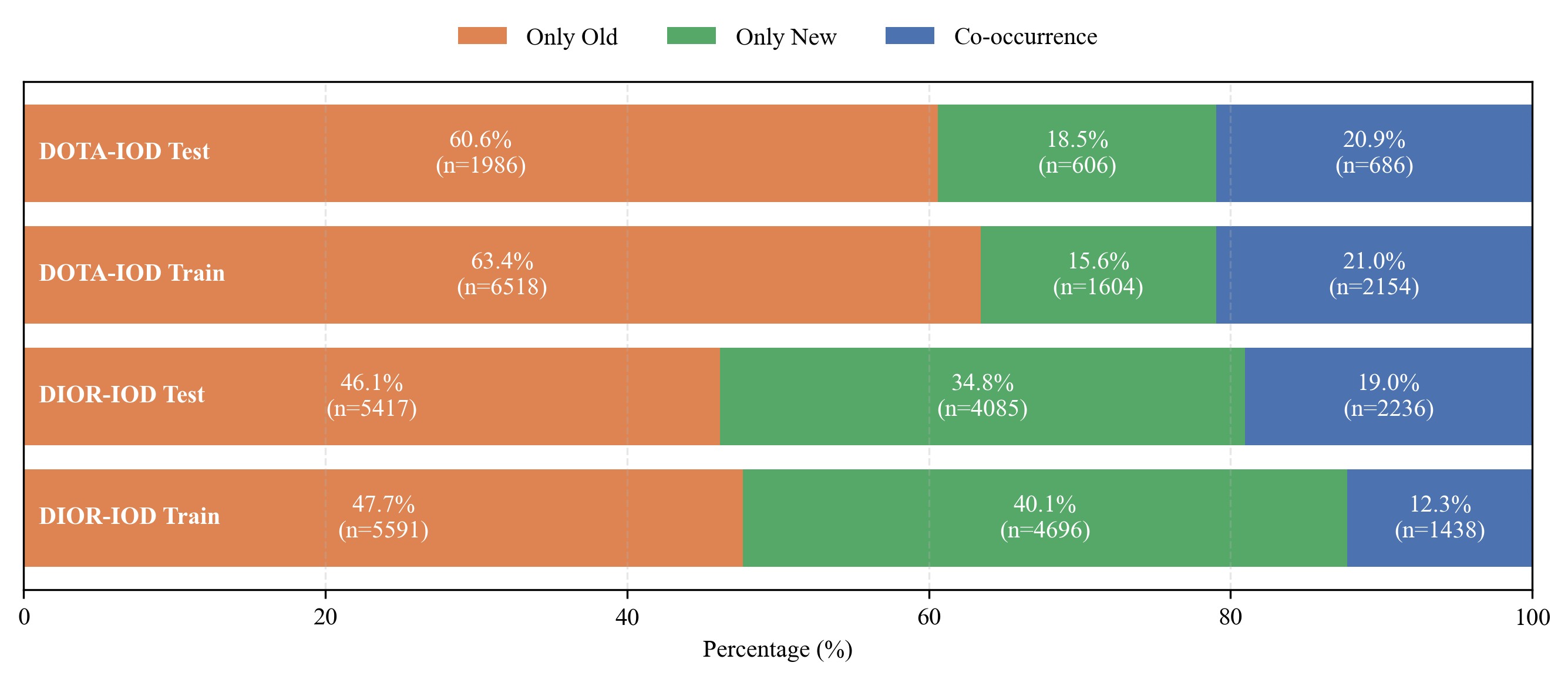}
    \caption{Distribution of Old–New Category Co-occurrence in DIOR-IOD and DOTA-IOD.}
    \label{dataset}
\end{figure}

{Since only new categories are annotated during the incremental phase, objects belonging to old categories remain unlabeled in the new data. To quantitatively analyze the distribution of new and old categories in the DIOR-IOD and DOTA-IOD datasets, we divide the dataset into three mutually exclusive subsets: \textbf{Only Old} (images containing only old categories and not used in the current incremental phase), \textbf{Only New} (images containing only new categories and used for training), and \textbf{Co-occurrence} (images containing both old and new categories, where old-category objects are unlabeled).
Among the training data available in the incremental phase (i.e., Only New + Co-occurrence), the frequency of co-occurrence is substantial, as shown in fig.~\ref{dataset}: it reaches 57.4\% in DOTA-IOD and 23.5\% in DIOR-IOD. This indicates that the model frequently encounters unlabeled old-category objects during incremental learning. If these objects are treated as background, they may introduce misleading supervision signals, leading to feature confusion and increased forgetting.}

{We adopt the standard COCO evaluation metrics~\citep{lin2014microsoft} as the primary performance criteria, including $\mathit{mAP}$, $\mathit{mAP}_{50}$, $\mathit{mAP}_{75}$, $\mathit{mAP}_{s}$, $\mathit{mAP}_{m}$, and $\mathit{mAP}_{l}$. In addition, we separately report the detection performance on previous task categories ($\mathit{mAP}^{\mathit{P}}$) and current task categories ($\mathit{mAP}^{\mathit{C}}$) to comprehensively evaluate the balance between knowledge retention and new class learning in IOD     \citep{gupta2022ow,zohar2023prob,sun2024exploring,xi2025ow}.}

\begin{table}[t]
\centering
\caption{The details of the DIOR-IOD dataset.}
\label{tab:dior}
\begin{tabular}{clcc}
\hline
{Task ID} & {Category} & {Train instance} & {Test instance} \\
\hline
\multirow{10}{*}{Task 1}
 & Airplane          & 1,888 & 8,212 \\
 & Airport           & 662 & 666  \\
 & Bridge                  & 1,367 & 2,589  \\
 & Service-area      & 1,080    & 1,085   \\
 & Toll-station     & 610    & 688   \\
 & Harbor            & 2,364  & 3,105    \\
 & Overpass                   & 1330 & 1,782 \\
 & Ship                 & 27,351  & 35,186    \\
 & Trainstation      & 501   & 509   \\
 & Vehicle           & 13,725  & 26,640  \\
\hline
\multirow{10}{*}{Task 2}
 & Baseballfield           & 2,384  & 3,434  \\
 & Basketballcourt               & 1,077 & 2,146  \\
  & Chimney             & 649    & 1,031    \\
 & Dam             & 512    & 538    \\
 & Golffield       & 511 & 575    \\
 & Groundtrackfield          & 1,162  & 1,885   \\
  & Stadium           & 595  & 672  \\
 & Storagetank                & 3,042 & 23,361  \\
 & Tenniscourt             & 4,898    & 7,343    \\
 & Windmill        & 2,365  & 2,998    \\
\hline
\end{tabular}
\end{table}

\begin{table}[t]
\centering
\caption{The details of the DOTA-IOD dataset.}
\label{tab:dota}
\begin{tabular}{clcc}
\hline
{Task ID} & {Category} & {Train instance} & {Test instance} \\
\hline
\multirow{5}{*}{Task 1}
 & Small-vehicle          & 49,864 & 10,833 \\
 & Large-vehicle          & 36,243 & 9,257  \\
 & Plane                  & 16,426 & 5,298  \\
 & Baseball-diamond       & 823    & 403    \\
 & Ground-track-field     & 838    & 316    \\
\hline
\multirow{5}{*}{Task 2}
 & Helicopter             & 1,193  & 134    \\
 & Ship                   & 61,075 & 19,571 \\
 & Bridge                 & 3,634  & 896    \\
 & Soccer-ball-field      & 849    & 351    \\
 & Tennis-court           & 4,738  & 1,602  \\
\hline
\multirow{5}{*}{Task 3}
 & Storage-tank           & 9,639  & 5,025  \\
 & Harbor                 & 13,085 & 4,789  \\
 & Roundabout             & 800    & 299    \\
 & Basketball-court       & 1,119  & 297    \\
 & Swimming-pool          & 3,324  & 763    \\
\hline
\end{tabular}
\end{table}

\begin{table*}[t]
\centering
\caption{Comparison with state-of-the-art methods on DIOR-IOD under the two-step IOD setting, the second-best results are \underline{underlined}.}
\setlength{\tabcolsep}{4pt}
\begin{tabular}{c|c|cc|cccccccc}
\toprule
\multirow{2}{*}{Method} 
& \multirow{2}{*}{Baseline}
& \multicolumn{2}{c|}{Task 1} 
& \multicolumn{8}{c}{Task 2} \\ 
\cmidrule(lr){3-4} \cmidrule(lr){5-12}
& 
& $\mathit{mAP}^{\mathit{C}}$
& $\mathit{mAP}^{\mathit{C}}_{50}$
& $\mathit{mAP}^{\mathit{A}}$
& $\mathit{mAP}^{\mathit{A}}_{50}$
& $\mathit{mAP}^{\mathit{A}}_{75}$
& $\mathit{mAP}^{\mathit{P}}$
& $\mathit{mAP}^{\mathit{P}}_{50}$ 
&$\mathit{mAP}^{\mathit{A}}_{s}$
&$\mathit{mAP}^{\mathit{A}}_{m}$
&$\mathit{mAP}^{\mathit{A}}_{l}$\\ 
\midrule
Fine-tuning & Grounding DINO & 45.6 & 72.7    & 22.6 & 34.6 & 24.4 & 1.3 & 3.6 & 4.9 & 17.8 & 32.6 \\
\midrule
MD-DETR     & Deformable DETR & 38.8 & 62.6  & 19.9 & 30.9 & 21.1 & 22.7 & 36.9 & 2.6 & 15.7 & 34.8\\
CL-DETR     & Deformable DETR & 34.8 & 54.4  & 31.5 & 43.9 & 34.5 & 17.5 & 28.5 & 5.2 & 23.5 & 46.6\\
\midrule
ERD         & Grounding DINO & 45.6 & 72.7   &31.4 & 45.3 & 32.8 & 8.6 & 14.7 & 8.1 & 24.5 & 41.8\\
GCD         & Grounding DINO & 45.6 & 72.7   &\underline{46.9} & \underline{71.2} & \underline{51.4} & \underline{39.4} & \underline{64.4} & \underline{15.1} & \underline{39.3} & \underline{66.9}\\
Ours        & Grounding DINO 
& 45.6 & 72.7  & \textbf{48.6\rlap{$_{\textcolor{red}{\uparrow 1.7}}$}} 
& \textbf{72.7\rlap{$_{\textcolor{red}{\uparrow 1.5}}$}} 
& \textbf{53.3\rlap{$_{\textcolor{red}{\uparrow 1.9}}$}} 
& \textbf{40.7\rlap{$_{\textcolor{red}{\uparrow 1.3}}$}} 
& \textbf{65.8\rlap{$_{\textcolor{red}{\uparrow 1.4}}$}} 
& \textbf{16.5\rlap{$_{\textcolor{red}{\uparrow 1.4}}$}}
& \textbf{41.4\rlap{$_{\textcolor{red}{\uparrow 2.1}}$}}
& \textbf{69.0\rlap{$_{\textcolor{red}{\uparrow 2.1}}$}}\\
\hline
\end{tabular}
\label{tab:dior_sota}
\end{table*}

\begin{table*}[htbp]
  \centering
  \caption{Comparison with state-of-the-art methods on DIOR-IOD under the multi-step IOD setting, the second-best results are underlined.}
  \label{tab:dior_multi_step}
  \resizebox{\textwidth}{!}{%
    \begin{tabular}{l| cc | cc | cc | cccccc}
      \toprule
      \multirow{2}{*}{Method} & \multicolumn{2}{c|}{\textit{Task 1}} & \multicolumn{2}{c|}{\textit{Task 2}} & \multicolumn{2}{c|}{\textit{Task 3}} & \multicolumn{6}{c}{\textit{Task 4}} \\
      \cmidrule(lr){2-3} \cmidrule(lr){4-5} \cmidrule(lr){6-7} \cmidrule(lr){8-13}
      & $mAP^A$ & $mAP^A_{50}$ & $mAP^A$ & $mAP^A_{50}$ & $mAP^A$ & $mAP^A_{50}$ & $mAP^A$ & $mAP^A_{50}$ & $mAP^P$ & $mAP^A_s$ & $mAP^A_m$ & $mAP^A_l$ \\
      \midrule
      MD-DETR & 43.0 & 61.5 & 34.4 & 45.8 & 31.1 & 43.0 & 19.1 & 27.4 & 19.0 & 3.3 & 14.9 & 31.3 \\
      CL-DETR & 42.2 & 60.7 & 36.0 & 48.8 & 30.9 & 43.0 & 22.1 & 31.7 & 22.6 & 3.4 & 15.6 & 35.7 \\
      ERD     & 44.0 & 69.0 & 29.0 & 38.2 & 14.3 & 21.5 & 18.1 & 28.9 & 11.5 & 9.1 & 19.4 & 27.2 \\
      GCD     & 44.0 & 69.0 & \underline{42.1} & \underline{59.1} & \underline{37.5} & \textbf{55.1} & \underline{34.5} & \underline{52.7} & \underline{33.6} & \underline{10.9} & \underline{31.2} & \underline{50.1} \\
      Ours    & 44.0 & 69.0 & \textbf{43.0\rlap{$_{\textcolor{red}{\uparrow 0.9}}$}} & \textbf{60.4\rlap{$_{\textcolor{red}{\uparrow 1.3}}$}} & \textbf{37.6\rlap{$_{\textcolor{red}{\uparrow 0.1}}$}} & {54.9\rlap{$_{\textcolor{green}{\downarrow 0.2}}$}} & \textbf{36.9\rlap{$_{\textcolor{red}{\uparrow 2.4}}$}} & \textbf{55.8\rlap{$_{\textcolor{red}{\uparrow 3.1}}$}} & \textbf{37.7\rlap{$_{\textcolor{red}{\uparrow 4.1}}$}} & \textbf{11.1\rlap{$_{\textcolor{red}{\uparrow 0.2}}$}} & \textbf{31.9\rlap{$_{\textcolor{red}{\uparrow 0.7}}$}} & \textbf{52.4\rlap{$_{\textcolor{red}{\uparrow 2.3}}$}} \\
      \bottomrule
    \end{tabular}%
  }
\end{table*}

\begin{table*}[t]
\centering
\caption{Comparison with state-of-the-art methods on DOTA-IOD under the multi-step IOD setting, the second-best results are \underline{underlined}.}
\small
\setlength{\tabcolsep}{3pt}
\begin{tabular}{c|cc|cccc|cccccccc}
\toprule
\multirow{2}{*}{Method} 
& \multicolumn{2}{c|}{Task 1} 
& \multicolumn{4}{c|}{Task 2} 
& \multicolumn{8}{c}{Task 3} 
\\ 
\cmidrule(lr){2-3} 
\cmidrule(lr){4-7} 
\cmidrule(lr){8-15}
& $\mathit{mAP}^{\mathit{C}}$
& $\mathit{mAP}^{\mathit{C}}_{50}$
& $\mathit{mAP}^{\mathit{A}}$
& $\mathit{mAP}^{\mathit{A}}_{50}$
& $\mathit{mAP}^{\mathit{P}}$
& $\mathit{mAP}^{\mathit{P}}_{50}$\hspace{0.6em}
& $\mathit{mAP}^{\mathit{A}}$
& $\mathit{mAP}^{\mathit{A}}_{50}$
& $\mathit{mAP}^{\mathit{A}}_{75}$
& $\mathit{mAP}^{\mathit{P}}$
& $\mathit{mAP}^{\mathit{P}}_{50}$ 
&$\mathit{mAP}^{\mathit{A}}_{s}$
&$\mathit{mAP}^{\mathit{A}}_{m}$
&$\mathit{mAP}^{\mathit{A}}_{l}$
\\ 
\midrule
Fine-tuning 
& 50.9 & 83.2 
& 21.7 & 37.2 & 12.5 & 22.5 
& 6.9 & 14.9 & 5.5 & 1.5 & 3.3 & 3.1 & 8.8 & 10.1\\
\midrule
MD-DETR     
& 38.8 & 62.6 
& 17.5 & 30.9 & 0.7 & 1.6 
& 8.4& 14.2 & 8.6 & 9.9 & 15.7 & 2.7 & 7.7 & 15.6\\
CL-DETR     
& 36.8 & 61.2 
& 27.3 & 43.1 & 23.4 & 39.2
& 17.9& 29.1 & 18.5 & 16.2 & 24.5 & 4.8 & 15.9 & 29.2\\
\midrule
ERD         
& 50.9 & 83.2
& 26.3 & 27.2 & 4.46 & 8.3 
& 13.1  & 23.6 & 13.1 & 0.7 & 1.5 & 5.9 & 15.5 & 18.7\\

GCD         
& 50.9 & 83.2
& \underline{42.1} & \underline{67.3} & \underline{47.5} & \underline{75.6} 
& \underline{37.3} & \underline{64.3} &\underline{40.9} & \underline{41.3} & \underline{62.1} & \textbf{20.6} & \underline{41.6} & \underline{45.2}\\

Ours        
& 50.9 & 83.2

& \textbf{42.4\rlap{$_{\textcolor{red}{\uparrow 0.3}}$}} 
& \textbf{67.8\rlap{$_{\textcolor{red}{\uparrow 0.5}}$}} 
& \textbf{49.1\rlap{$_{\textcolor{red}{\uparrow 1.6}}$}} 
& \textbf{76.6\rlap{$_{\textcolor{red}{\uparrow 1.0}}$}}
& \textbf{39.4\rlap{$_{\textcolor{red}{\uparrow 2.1}}$}} 
& \textbf{65.5\rlap{$_{\textcolor{red}{\uparrow 1.2}}$}} 
& \textbf{44.1\rlap{$_{\textcolor{red}{\uparrow 3.2}}$}} 
& \textbf{42.7\rlap{$_{\textcolor{red}{\uparrow 1.4}}$}} 
& \textbf{63.9\rlap{$_{\textcolor{red}{\uparrow 1.8}}$}}
&{20.3\rlap{$_{\textcolor{green}{\downarrow 0.3}}$}}
& \textbf{42.8\rlap{$_{\textcolor{red}{\uparrow 0.8}}$}}
& \textbf{48.4\rlap{$_{\textcolor{red}{\uparrow 3.2}}$}}
\\
\bottomrule
\end{tabular}
\label{tab:dota_sota}
\end{table*}

\subsection{Comparison with the State-of-the-art Methods}
To assess the performance of STAR-IOD in remote sensing scenarios, we performed comparative experiments on the DIOR-IOD and DOTA-IOD datasets against the Deformable DETR-based CL-DETR \citep{liu2023continual} and MD-DETR \citep{bhatt2024preventing}, as well as the Grounding DINO-based ERD \citep{feng2022overcoming} and GCD \citep{wang2025gcd}.

\noindent\textbf{Comparison Experiments on the DIOR-IOD under two-step setting:} As presented in Table~\ref{tab:dior_sota}, we conducted a comprehensive comparison between the proposed method and existing SOTA approaches on the DIOR-IOD dataset under the two-step setting. The quantitative results demonstrate that our method achieves superior performance across all evaluation metrics. Specifically, the direct Fine-tuning baseline suffers from severe catastrophic forgetting in Task 2, where the retention of previously learned classes drastically collapses to 1.3\% $\mathit{mAP}^P$. 
Compared to methods based on Deformable DETR (i.e., MD-DETR and CL-DETR), approaches utilizing Grounding DINO exhibit higher overall performance, benefiting from their robust multi-modal representations. However, despite being built upon Grounding DINO, the ERD method still struggles with retaining old knowledge, yielding a low $\mathit{mAP}^P$ of only 8.6\%.  In comparison to GCD, the second-best performing method, our approach achieves consistent improvements across all metrics. Specifically, in terms of the comprehensive metric for all classes, we achieve 48.6\% $\mathit{mAP}^A$, outperforming GCD by 1.7\% percentage points. Furthermore, on the critical metric for preventing forgetting, our method reaches 40.7\% $\mathit{mAP}^P$, representing a gain of 1.3 percentage points. Additionally, with a slight decrease of only 3.3\% $mAP$ compared to the joint-training baseline, STAR-IOD demonstrates superior anti-forgetting performance in RS-IOD. {Furthermore, STAR-IOD consistently outperforms competing methods across all three object scales. Compared to the second-best approach, GCD, it yields notable gains of +1.4\%, +2.1\%, and +2.1\% in $\mathit{mAP}_s^A$, $\mathit{mAP}_m^A$, and $\mathit{mAP}_l^A$, respectively. These results indicate that, facilitated by the scale subspace decoupling mechanism within the STD, STAR-IOD effectively addresses intra-class scale variations. Consequently, the overall performance enhancement is not skewed towards any single scale, but rather reflects a highly balanced optimization across diverse object sizes.}

\begin{figure*}
    \centering
    \includegraphics[width=0.8\linewidth]{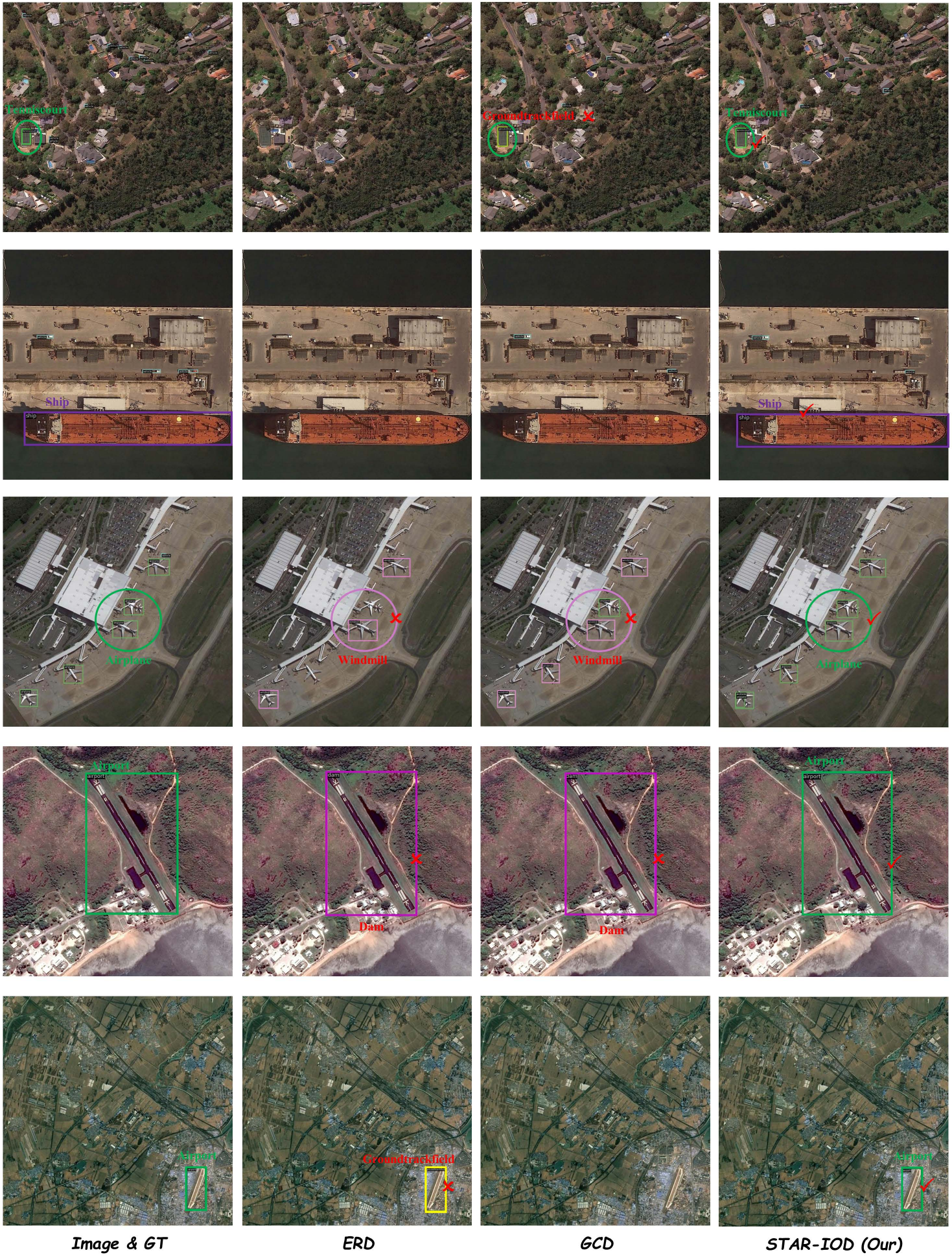}
    \caption{Qualitative comparison of detection results on the DIOR-IOD dataset across different methods.}
    \label{qualitative_dior}
\end{figure*}

We present qualitative comparisons of the detection results produced by ERD, GCD, and the proposed STAR-IOD on the DIOR-IOD dataset, as shown in Fig.~\ref{qualitative_dior}. It can be observed that ERD and GCD struggle to preserve performance on previously learned categories, exhibiting two representative types of errors. First, missed detections frequently occur as a consequence of old knowledge degradation. For example, in the second row, both ERD and GCD fail to detect the ship, indicating catastrophic forgetting during the incremental learning process. Second, as illustrated in rows 3–5, the competing methods also suffer from pronounced misclassification. {Airplanes} are incorrectly recognized as {Windmills}. This issue arises because the feature representations of old classes are disrupted by newly introduced classes during incremental updates. Such interference leads to semantic drift, which distorts the original decision boundaries and results in substantial semantic confusion between visually similar objects.

\noindent\textbf{Comparison Experiments on the DIOR-IOD under multi-step setting:} {To further validate the generalization ability and robustness of the proposed method under longer incremental sequences and more complex semantic interference, we construct a more challenging multi-step incremental learning setting on the DIOR-IOD dataset and conduct comparisons with state-of-the-art methods. The quantitative results are reported in Table~\ref{tab:dior_multi_step}. As the incremental tasks progress, methods such as MD-DETR, CL-DETR, and ERD suffer from severe catastrophic forgetting, whereas our method demonstrates strong stability and continual learning capability. At the final stage (Task 4), after multiple rounds of learning new classes, our method achieves the best performance across all metrics. Compared with the competing method GCD, our method improves the overall accuracy, achieving gains of 2.4\% in $\mathit{mAP}^{A}$ and 3.1\% in $\mathit{mAP}_{50}^{A}$. In terms of old-class performance, our method achieves an $\mathit{mAP}^{P}$ of 37.7\% in Task 4, outperforming GCD by a substantial margin of 4.1\%. This clearly demonstrates that our method can more effectively anchor and preserve previously learned knowledge under the challenges of imbalanced category distribution and complex background interference in remote sensing imagery, thereby achieving a better balance between acquiring new knowledge and retaining old knowledge. Furthermore, to address the inherent challenge of scale variation in remote sensing imagery, our method also exhibits stronger adaptability in multi-scale object detection. On the multi-scale metrics $\mathit{mAP}_{s}^{A}$, $\mathit{mAP}_{m}^{A}$, and $\mathit{mAP}_{l}^{A}$ in Task 4, our method consistently surpasses all competing baselines, achieving improvements of 0.2\%, 0.7\%, and 2.3\% over GCD, respectively.}

\noindent\textbf{Comparison Experiments on the DOTA-IOD: }As shown in Table~\ref{tab:dota_sota}, we compare the performance of STAR-IOD with current SOTA methods on the DOTA-IOD dataset under the multi-step setting. 
Compared to the two-step setting, multi-step IOD poses significantly more stringent challenges to the model’s ability to retain previously learned knowledge. 
Specifically, Fine-tuning and ERD suffer from severe catastrophic forgetting in Task 3, with their performance on previous classes collapsing to 1.5\% and 0.7\% $\mathit{mAP}^P$, respectively. 
In contrast, STAR-IOD exhibits remarkable robustness, with its performance advantage becoming more pronounced as the incremental tasks increase. 
Although STAR-IOD achieves only a marginal improvement of 0.3\% in $\mathit{mAP}^A$ over the second-best method (GCD) in Task 2, this performance gap widens substantially to 2.1\% in Task 3.
Crucially, we achieve a significant gain of 3.2\% $\mathit{mAP}^A_{75}$ in the high-precision localization metric. 
This superior performance indicates that STAR-IOD more effectively suppresses error accumulation than existing approaches, not only mitigating category forgetting but also maintaining bounding box regression accuracy during long-term incremental learning. 

{Beyond its robustness against catastrophic forgetting, the results in Table~\ref{tab:dota_sota} further validate the capability of STAR-IOD to handle significant scale variations. Compared with GCD, STAR-IOD achieves improvements of +0.8\% on medium-sized objects and a larger gain of +3.2\% on large objects. Meanwhile, a marginal decrease of 0.3\% is observed on small objects. This is mainly attributed to the characteristics of the DOTA-IOD dataset, which contains a large number of densely distributed small objects; in the absence of explicit scale partitioning, conventional methods tend to bias the optimization process toward learning representations for small objects. By contrast, STAR-IOD alleviates this bias by decoupling scale-specific feature subspaces, thereby leading to a more balanced optimization across object scales.}

\subsection{Ablation Study}
In this section, we perform comprehensive ablation studies to isolate and evaluate the specific contribution of each component within the STAR-IOD framework. In addition, we investigate the threshold setting in the CPG module and determine its optimal value through empirical evaluation.

\noindent\textbf{Ablation Study on Different Modules for STAR-IOD: }
As shown in Table~\ref{abltation}, directly fine-tuning Grounding DINO leads to severe catastrophic forgetting, causing a drastic drop in detection performance on old categories. The detection accuracy for old classes falls to only 1.5\% $mAP$, indicating that the model almost completely loses its ability to represent previously learned knowledge during incremental learning. 
CRD compels the student model to simultaneously mimic the teacher’s classification score distributions and bounding box regression outputs at the level of semantically corresponding queries. By aligning both classification responses and localization predictions, the student is encouraged to inherit the old-class discriminative capability and spatial localization knowledge embedded in the teacher while learning new categories, thereby alleviating catastrophic forgetting to some extent in the incremental learning process. Nevertheless, the preservation of old knowledge achieved by CRD remains limited, with the detection performance on old categories improving only to 8.3\% $mAP$. This indicates that relying solely on response-level distillation is insufficient to maintain the discriminative structure of old classes, and the model still suffers from evident forgetting.
Incorporating CPG significantly improves the model’s performance on old categories, increasing the detection accuracy of old classes to 30.4\% $\mathit{mAP}^P$. CPG compensates for missing annotations of old categories by generating high-quality pseudo-labels. Combining CRD with CPG further enhances the model’s stability, increasing $\mathit{mAP}^P$ to 40.7\%. This suggests that response-level distillation becomes more effective at preserving old knowledge when supported by pseudo-label supervision. However, despite the improvement on old categories, the detection performance on new classes declines, with $\mathit{mAP}^C$ dropping to 52.8\%, indicating an inherent trade-off between stability and plasticity. By aligning multi-scale topological structures, STD improves the detection performance on new categories to 57.4\% $\mathit{mAP}^C$ while maintaining the stability of old classes. Consequently, STAR-IOD achieves a favorable balance between stability and plasticity, reaching an overall detection accuracy of 48.6\% $\mathit{mAP}^A$.

\begin{table}[htbp]
\centering
\caption{Ablation study of different modules on the DIOR-IOD dataset.}
\renewcommand{\arraystretch}{1.5} 
\setlength{\tabcolsep}{1.5pt}       
\footnotesize  
\begin{tabular}{ccc|ccccccc}
\hline
\multicolumn{3}{c|}{Module} & \multicolumn{7}{c}{Task2} \\

CRD & CPG & STD & $mAP^{{A}}$ & $mAP_{50}^{{A}}$ & $mAP_{75}^{{A}}$ & $mAP^{{P}}$ & $mAP_{50}^{{P}}$ & $mAP^{{C}}$ & $mAP_{50}^{{C}}$ \\
\hline
 & & & 22.6 & 34.6 & 24.4 & 1.5 & 3.6 & 43.9 & 65.8 \\

$\checkmark$ & & & 31.2 & 44.3 & 33.8 & 8.3 & 13.7 & 54.2 & 74.9 \\
 & $\checkmark$ & &  43.5& 64.5 & 46.6 & 30.4 & 52.3 & 57.2 & 76.7 \\
$\checkmark$ & $\checkmark$ & & 46.4 & 69.9 & 49.9 & 40.7 & 65.6 & 52.8 & 74.4 \\
$\checkmark$ & $\checkmark$ & $\checkmark$ & 48.6 & 72.7 & 53.3 & 40.7 & 65.8 & 57.4 & 78.4 \\
\hline
\end{tabular}
\label{abltation}
\end{table}

\noindent\textbf{Ablation Study on CPG: }{To validate the effectiveness and adaptability of the CPG module in the RS-IOD, we analyze the impact of $\delta_{min}$, and further conduct a comprehensive comparison with the traditional Fixed-threshold pseudo-labeling (ST) as well as two advanced adaptive pseudo-labeling methods, namely Consistent Teacher~\citep{wang2023consistent} and LPL~\citep{yoon2024enhancing}. As shown in Table~\ref{tab:cpg_ablation}, under the optimal setting, CPG outperforms all compared methods, demonstrating its clear advantage in RS-IOD.}

Specifically, when $\delta_{min}=0.1$, an excessive number of low-confidence noisy predictions are introduced into the candidate pool, causing the estimated optimal threshold to be biased toward the low-confidence cluster. As a result, a non-negligible number of FP are incorporated into the pseudo-label set, which in turn limits the overall detection performance. As $\delta_{min}$ increases to 0.2 and 0.3, CPG achieves steady improvements across both overall and old-class metrics. In particular, $\delta_{min}=0.3$ yields the best performance on most evaluation metrics, indicating that a moderately higher candidate threshold effectively suppresses noisy pseudo-labels while retaining sufficient high-quality samples for reliable threshold estimation.

In contrast, the ST baseline is highly sensitive to the choice of threshold, exhibiting noticeable performance fluctuations as the threshold varies. Even under its best configuration ($\delta=0.4$), ST remains clearly inferior to CPG in terms of both overall accuracy and old-class retention. These results suggest that a fixed threshold is inadequate for handling the class- and stage-dependent confidence distribution shifts in incremental learning, whereas CPG dynamically adapts category-specific thresholds in a data-driven manner, producing more reliable pseudo-labels and effectively mitigating catastrophic forgetting.

{More importantly, in comparisons with advanced dynamic thresholding methods, CPG also demonstrates superiority. Consistent Teacher serves as a representative adaptive-threshold baseline. It models the distribution of prediction scores using a Gaussian Mixture Model (GMM), enabling automatic threshold selection for distinguishing foreground and background. This approach provides a robust alternative to manually defined thresholds. Low-confidence Pseudo Label Distillation focuses on mining hard-to-detect objects. LPL first extracts high-confidence pseudo-labels, then identifies low-confidence pseudo-labels from proposals with minimal overlap with the high-confidence set. These low-confidence proposals are further refined by removing background scores and applying L1 normalization to foreground scores, enhancing category-discriminative features. Experimental results show that under the optimal setting of $\delta_{\min}=0.3$, CPG achieves 46.5\% $mAP^A$ and 41.1\% $mAP^P$. Compared to Consistent Teacher, which obtains 46.1\% $mAP^A$ and an $mAP^P$ of 40.2\%, CPG achieves superior performance. Consistent Teacher assumes that the confidence scores follow a Gaussian distribution; however, in the RS-IOD scenario, this assumption is often violated due to noise, occlusion, and class imbalance. In contrast, CPG does not rely on any predefined distribution assumption, making it more suitable for the RS-IOD scenario and leading to more accurate and robust threshold estimation. Moreover, compared to LPL, CPG avoids the multi-stage pseudo-label mining process and produces higher-quality pseudo-labels, achieving an $mAP^P_{50}$ of 65.6\%, which exceeds the 61.9\% obtained by LPL.}

\begin{table}[htbp]
  \centering
  \caption{Ablation study of the CPG module and performance comparison with different fixed and dynamic thresholding baselines. The best results are highlighted in \textbf{bold}.}
  \renewcommand{\arraystretch}{1.0} 
  \setlength{\tabcolsep}{3pt}      
  
  \begin{tabular}{lccccc}
    \toprule
    \multicolumn{1}{c}{Module} & $mAP^{A}$ & $mAP^{A}_{50}$ & $mAP^{A}_{75}$ & $mAP^{P}$ & $mAP^{P}_{50}$ \\
    \midrule
    CPG ($\delta_{min}=0.1$) & 44.1 & 67.6 & 47.4 & 40.5 & 65.4 \\
    CPG ($\delta_{min}=0.2$) & 46.4 & \textbf{69.9} & 49.9 & 40.7 & \textbf{65.6} \\
    CPG ($\delta_{min}=0.3$) & \textbf{46.5} & \textbf{69.9} & \textbf{50.3} & \textbf{41.1} & \textbf{65.6} \\
    \midrule
    ST ($\delta=0.2$) & 43.6 & 67.3 & 46.9 & 38.7 & 63.9 \\
    ST ($\delta=0.3$) & 44.9 & 68.5 & 48.4 & 40.1 & 65.3 \\
    ST ($\delta=0.4$) & 45.3 & 68.8 & 49.3 & 39.7 & 64.5 \\
    \midrule
    Consistent Teacher & 46.1 & 69.6 & 49.6 & 40.2 & 64.8 \\
    LPL & 45.5 & 68.9 & 49.1 & 40.0 & 61.9 \\
    \bottomrule
  \end{tabular}
  \label{tab:cpg_ablation}
\end{table}
\begin{figure}
    \centering
    \includegraphics[width=0.9\linewidth]{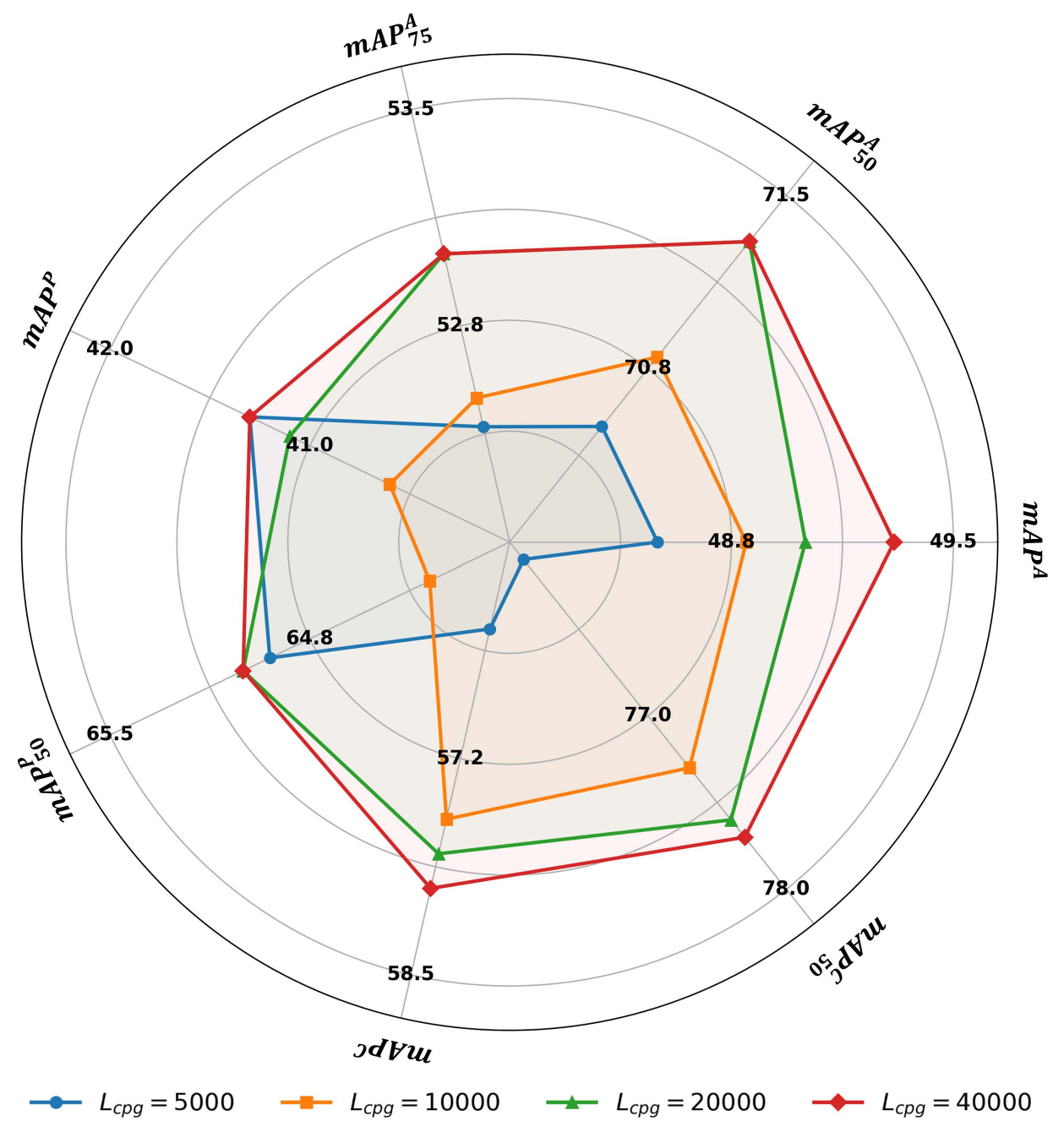}
    \caption{Impact of different $L_{cpg}$ settings on detection performance on the DOTA-IOD dataset.}
    \label{lcpg}
\end{figure}

\noindent
{\textbf{Effectiveness of Scale-adaptive Instance Partitioning:} 
To further validate the necessity of Scale-adaptive Instance Partitioning (SIP), we construct a variant where this component is removed. In this setting, all instances of the same class, regardless of their scale differences, are used to compute a single prototype, followed by topology alignment. Without SIP, scale-variant instances are forced to contribute to a unified class prototype, which may bias the prototype toward the dominant scale group and further distort the resulting relation matrix.}

{As shown in Table~\ref{tab:std}, removing SIP leads to a clear performance degradation across all evaluation metrics. Specifically, the overall $\mathit{mAP}^{\mathit{A}}$ drops from $39.4\%$ to $37.3\%$, indicating a notable decline in detection performance. In contrast, the proposed SIP consistently improves all evaluation metrics. More importantly, it improves the overall $\mathit{mAP}^{\mathit{A}}$ by $+2.1\%$, with gains of $+0.7\%$, $+0.8\%$, and $+3.2\%$ on small, medium, and large objects, respectively. These results indicate that SIP partitions instances into scale-consistent groups, enabling more compact and coherent feature distributions. This scale decoupling effectively prevents the formation of biased relation matrices during incremental learning, thereby preserving the structural stability of learned representations and facilitating more precise topological knowledge transfer.}

\begin{table}[htbp]
\centering
\caption{Ablation study of Scale-adaptive Instance Partitioning.}
\setlength{\tabcolsep}{5pt}
\label{tab:std}
\begin{tabular}{cccccc}
\hline
     & $\mathit{mAP}^{A}$ & $\mathit{mAP}^{A}_{50}$ & $\mathit{mAP}_{s}$ & $\mathit{mAP}_{m}$ & $\mathit{mAP}_{l}$ \\
\hline
w/o SIP  & 37.3 & 64.1 & 19.6 & 41.6 & 45.2 \\
w/ SIP  &\textbf{39.4\rlap{$_{\textcolor{red}{\uparrow 2.1}}$}}
     &\textbf{65.5\rlap{$_{\textcolor{red}{\uparrow 1.4}}$}}
     &\textbf{20.3\rlap{$_{\textcolor{red}{\uparrow 0.7}}$}}
     &\textbf{42.4\rlap{$_{\textcolor{red}{\uparrow 0.8}}$}}
     &\textbf{48.4\rlap{$_{\textcolor{red}{\uparrow 3.2}}$}}  \\
     
\hline
\end{tabular}
\end{table}

\noindent
{\textbf{Analysis of loss weight $\lambda_1$: }We evaluate the hyperparameter $\lambda_1 \in \{1,2,3,4\}$ to determine its optimal setting. As shown in Table \ref{tab:lambda}, the results indicate that $\lambda_1 = 3$ achieves the best performance on $\mathit{mAP}^A$, $\mathit{mAP}_{50}^A$, and $\mathit{mAP}_{75}^A$. While $\lambda_1 = 1$ better preserves the performance of old classes, this gain comes at the expense of plasticity, leading to a 2.2\% decrease in $\mathit{mAP}^C$. Consequently, the overall performance improves by 0.5\%. This demonstrates that $\lambda_1 = 3$ provides a better balance between stability and plasticity, effectively mitigating catastrophic forgetting while accommodating new classes.}

\noindent
{\textbf{Analysis of scale partition thresholds $(\tau_s, \tau_m)$: }
To evaluate the impact of scale partitioning on model performance, we conduct an ablation study on the thresholds $\tau_s$ and $\tau_m$. As shown in Table~\ref{tab:tau}, the setting $(\tau_s, \tau_m) = (1024, 9216)$ achieves the optimal performance among the evaluated configurations. It obtains the highest accuracy in five out of six evaluation metrics, notably achieving 49.4\% on $\mathit{mAP}^A$. The setting $(1024, 9216)$ achieves a better balance across different object scales. Although it leads to a slight decrease in small-object performance, the improvements on medium and large objects result in higher overall performance. These results indicate that this threshold setting can alleviate intra-class scale variation in remote sensing scenarios. We observe that even the worst-performing scale partitioning threshold configuration $(2048, 8192)$ outperforms the comparative method GCD. Therefore, the scale-adaptive subspace partitioning mechanism effectively mitigates feature distribution shift caused by intra-class scale variations in remote sensing scenarios, thereby maintaining the stability and accuracy of prior knowledge distillation.}

{Since STD performs topology alignment by modeling inter-class relations independently within each scale-specific subspace, the effectiveness of scale partitioning directly governs the stability of feature distributions and the statistical adequacy within each subspace. The configuration $(1024,9216)$ yields a more balanced partition across scales, ensuring that each subspace maintains representative coverage without being dominated by a particular scale. This leads to more stable prototype estimation and more reliable relation modeling.}

\begin{table}[htbp]
\centering
\caption{Performance Evaluation under Different $\lambda_1$ Settings}
\label{tab:lambda}
\setlength{\tabcolsep}{6pt}
\begin{tabular}{cccccc}
\toprule
$\lambda_1$ & $\mathit{mAP}^{A}$ & $\mathit{mAP}^{A}_{50}$ & $\mathit{mAP}^{A}_{75}$ & $\mathit{mAP}^{P}$ & $\mathit{mAP}^{C}$ \\
\midrule
1 & 48.9 & 71.2 & 52.7 & \textbf{42.1} & 56.4 \\
2 & 47.9 & 69.8 & 51.8 & 40.9 & 55.7 \\
3 & \textbf{49.4} & \textbf{71.6} & \textbf{53.4} & 41.2 & \textbf{58.6} \\
4 & 48.9 & 71.1 & 52.9 & 42.0 & 56.6 \\
\bottomrule
\end{tabular}
\end{table}

\begin{table}[htbp]
\centering
\setlength{\tabcolsep}{3pt}
\caption{Ablation study on the scale partition thresholds.}
\label{tab:tau}
\begin{tabular}{ccccccc}
\toprule
$(\tau_s, \tau_m)$ & $\mathit{mAP}^{A}$ & $\mathit{mAP}^{A}_{50}$ & $\mathit{mAP}^{A}_{75}$ & $\mathit{mAP}_{s}$ & $\mathit{mAP}_{m}$ & $\mathit{mAP}_{l}$ \\
\midrule
(512, 4096)  & 48.9 & 71.3 & 52.8 & 16.8 & 41.2 & 68.1 \\
(1024, 4096) & 48.8 & 70.7 & 52.7 & 16.3 & 40.9 & 67.8 \\
(1024, 9216) & \textbf{49.4} & \textbf{71.6} & \textbf{53.4} & 16.7 & \textbf{41.6} & \textbf{68.7} \\
(2048, 8192) & 48.4 & 70.6 & 52.1 & \textbf{17.2} & 41.1 & 67.4 \\
(2048, 9216) & 48.9 & 71.1 & 52.9 &16.7      &40.7      & 68.3     \\
\bottomrule
\end{tabular}
\end{table}
\noindent\textbf{Ablation Study on CPG $L_{cpg}$: }
We further analyze the effect of the  score bank size $L_{cpg}$ in the CPG module, as illustrated in Fig.~\ref{lcpg}. As $L_{cpg}$ increases from 5000 to 20000, the detection performance shows a consistent improvement, particularly in overall accuracy and high-precision localization, indicating that a larger confidence bank enables more reliable clustering-based threshold estimation. When $L_{cpg}$ is too small, the limited samples are insufficient to capture the confidence distribution, leading to suboptimal pseudo-label filtering. Although slightly better results can be observed when further increasing $L_{cpg}$ to 40000, the performance gains become marginal. Therefore, we adopt $L_{cpg}=20000$ in our experiments.

\noindent\textbf{Old and New Class Results in Multi-Step Setting:}
Table~\ref{class} presents the detailed quantitative performance for each category on the DOTA-IOD dataset. Our method achieves a comprehensive $mAP$ of 39.4\% across all 15 categories. Specifically, regarding the old categories (the first 10 classes), the model demonstrates robust retention of prior knowledge, attaining high precision on classes such as {Tennis-court} (91.6\% ${AP}_{50}$), {Plane} (81.3\% ${AP}_{50}$), and {Ship} (82.7\% ${AP}_{50}$). This indicates that STAR-IOD effectively mitigates the issue of catastrophic forgetting in remote sensing object detection. Furthermore, concerning the new categories (the last 5 classes), our method exhibits favorable plasticity, yielding competitive results on {Basketball-court} (63.5\% ${AP}_{50}$) and {Storage-tank} (72.5\% ${mAP}_{50}$). These results confirm that our proposed framework strikes an optimal balance between stability on previous tasks and plasticity on new tasks.

\begin{figure}
    \centering
    \includegraphics[width=1\linewidth]{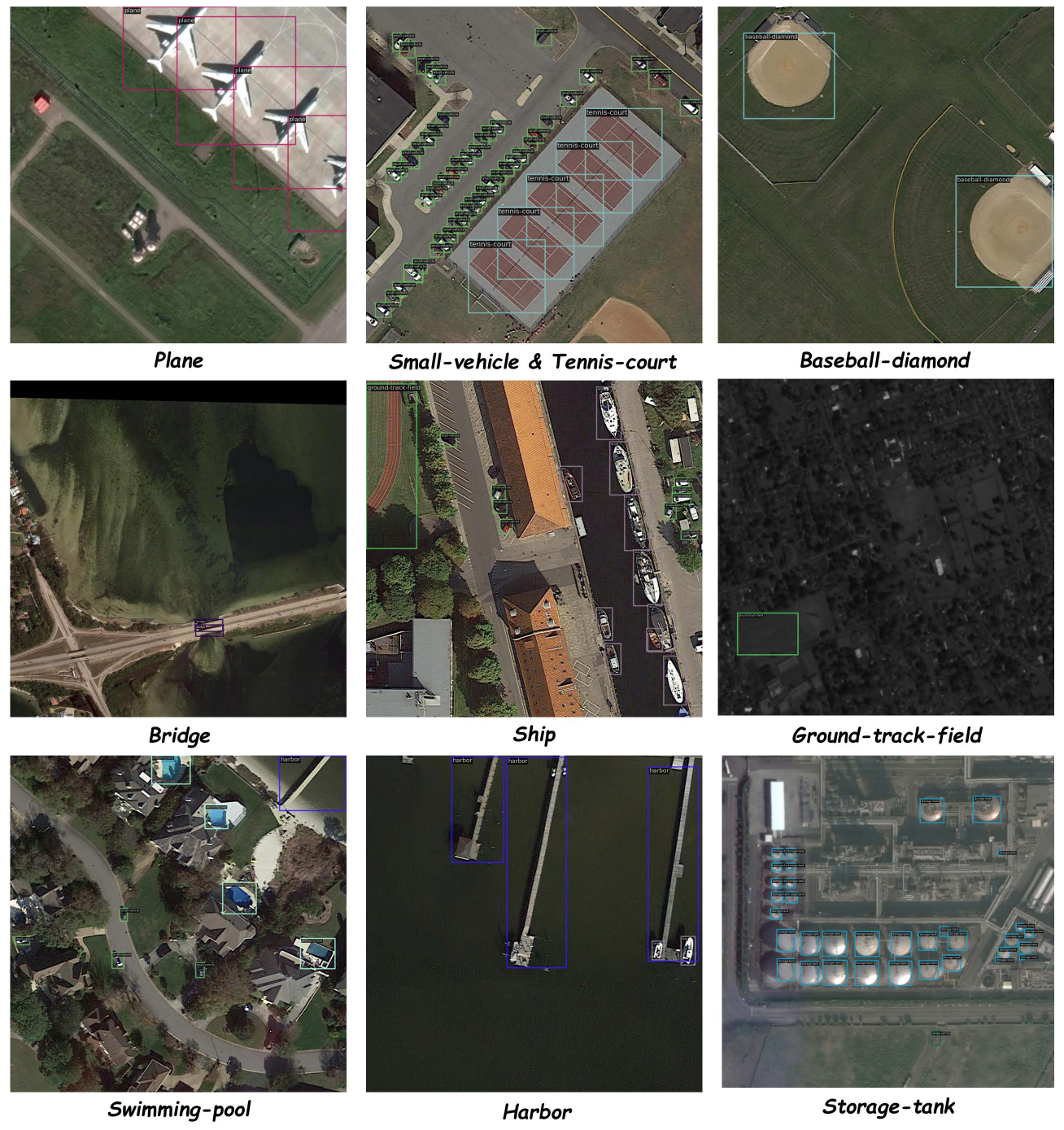}
    \caption{Detection visualization results of old and new classes on the DOTA-IOD dataset. The first two rows show old classes, while the third row shows new classes.
}
    \label{old_new}
\end{figure}

In addition, we visualize the qualitative detection results of both old and new categories under a multi-step incremental setting on DOTA-IOD, as presented in Fig.~\ref{old_new}. As illustrated in the figure, categories involved in the Task 1 ($T_1$), such as {Plane}, {Small-vehicle}, and {Baseball-diamond}, maintain precise detection results. Notably, even under drastic imaging environmental changes affecting the {Ground-track-field} category, STAR-IOD retains excellent detection capability for old categories. This finding indicates that STAR-IOD can, to a certain extent, achieve cross-domain incremental detection tasks. 

Furthermore, categories from the subsequent incremental stage $T_2$ incremental stage, specifically {Ship} and {Bridge}, are effectively transferred to the student model, ensuring that the student model preserves complete historical knowledge. Regarding the detection results for novel categories introduced in the latest stage, the model exhibits strong plasticity, accurately localizing and classifying these new objects. This verifies the effectiveness of our framework in acquiring new knowledge while maintaining stability on previous tasks.

\begin{table}[htbp]
  \centering
  \caption{Per-class detection accuracy (\%) on the DOTA-IOD dataset.}
  \resizebox{\linewidth}{!}{
  \begin{tabular}{clcccccc}
    \toprule
    Task & Category & $AP$ & ${AP}_{50}$ & ${AP}_{75}$ & ${AP}_s$ & ${AP}_m$ & ${AP}_l$\\
    \midrule
    
    \rowcolor{gray!15} 
     & Small-vehicle & 34.0 & 56.9 & 37.3 & 28.0 & 50.1 & - \\
    \rowcolor{gray!15} 
     & Large-vehicle & 55.2 & 78.9 & 66.6 & 32.2 & 60.3 & 62.1 \\
    \rowcolor{gray!15} 
     & Plane & 57.3 & 81.3 & 69.5 & 23.3 & 59.9 & 67.0 \\
    \rowcolor{gray!15} 
     & Baseball-diamond & 37.5 & 72.1 & 36.7 & 8.2 & 26.9 & 53.4 \\
    \rowcolor{gray!15} 
    \multirow{-5}{*}{Task 1} & Ground-track-field & 47.7 & 71.9 & 56.2 & 8.3 & 40.7 & 54.8 \\
    
    \rowcolor{green!10} 
     & Helicopter & 0.9 & 1.6 & 0.9 & 0.4 & 1.0 & 1.1 \\
    \rowcolor{green!10} 
     & Ship & 56.9 & 82.7 & 70.0 & 46.8 & 64.5 & 53.2 \\
    \rowcolor{green!10} 
     & Bridge & 20.9 & 45.3 & 16.1 & 19.8 & 23.0 & 22.1 \\
    \rowcolor{green!10} 
     & Soccer-ball-field & 35.6 & 57.0 & 40.1 & 9.1 & 23.5 & 44.3 \\
    \rowcolor{green!10} 
    \multirow{-5}{*}{Task 2} & Tennis-court & 81.4 & 91.6 & 89.3 & 37.7 & 71.1 & 90.3 \\
    

    \multirow{5}{*}{Task 3} & Storage-tank & 42.5 & 72.5 & 45.1 & 33.5 & 61.3 & 69.8 \\
     & Harbor & 41.0 & 76.9 & 40.2 & 28.9 & 45.1 & 43.7 \\
     & Roundabout & 32.3 & 62.9 & 30.4 & 14.8 & 43.1 & 14.9 \\
     & Basketball-court & 45.0 & 63.5 & 49.2 & 0.3 & 29.5 & 63.9 \\
     & Swimming-pool & 27.7 & 67.0 & 13.7 & 12.8 & 36.5 & 37.1 \\
    
    \midrule
    & {Average} & {39.4} & {65.5} & {44.1} & {20.3} & {42.4} & {48.4} \\
    \bottomrule
  \end{tabular}
  }
  \label{class}
\end{table}
\noindent\textbf{Forgetting Analysis of STAR-IOD: }
To quantitatively investigate the impact of catastrophic forgetting during the incremental learning phase, we present a detailed performance comparison between the $T_1$ and $T_2$ in Table~\ref{forget_analysis}.As observed, the model inevitably suffers from performance degradation across most base categories due to the stability-plasticity dilemma, where the optimization for new objectives leads to a drift in the feature space of previously learned classes. Specifically, categories with relatively low detection performance at the initial stage, such as {Harbor} and {Bridge}, exhibit the most severe forgetting, with ${AP}_{50}$ drops of $19.1\%$ and $11.5\%$, respectively. This phenomenon indicates that, during the knowledge distillation process, unreliable or weakly discriminative representations from the teacher are simultaneously transferred to the student, introducing noisy and ambiguous supervision that further degrades the student’s discriminative capability for these categories and exacerbates performance deterioration.

In contrast, categories with rigid geometric structures and distinct features demonstrated remarkable resilience. Notably, the Toll-station category maintained its performance with a negligible decline in ${AP}_{50}$, and Service-area experienced only a minor drop of $4.4\%$. The high retention rate for these classes indicates that our method effectively preserves strong discriminative features for distinct objects. 

\begin{table}[htbp]
  \centering
  \caption{Comparison of detection performance between Task 1 and Task 2 (in \%) on DIOR-IOD. The performance drops in Task 2 are marked with green arrows.}
  \label{forget_analysis}
  \resizebox{\linewidth}{!}{
  \begin{tabular}{lcccccc}
    \toprule
    \multirow{2}{*}{Category} & \multicolumn{3}{c}{Task 1} & \multicolumn{3}{c}{Task 2} \\
    \cmidrule(lr){2-4} \cmidrule(lr){5-7}
     & $AP$ & ${AP}_{50}$ & ${AP}_{75}$ & AP & ${AP}_{50}$ & ${AP}_{75}$ \\
    \midrule
    Airplane & 46.0 & 65.9 & 52.5 & 41.1\rlap{$_{\textcolor{green!65!black}{\downarrow 4.9}}$} & 52.9\rlap{$_{\textcolor{green!65!black}{\downarrow 13.0}}$} & 46.9\rlap{$_{\textcolor{green!65!black}{\downarrow 5.6}}$} \\
    Airport & 62.6 & 90.3 & 69.7 & 57.3\rlap{$_{\textcolor{green!65!black}{\downarrow 5.3}}$} & 84.6\rlap{$_{\textcolor{green!65!black}{\downarrow 5.7}}$} & 62.2\rlap{$_{\textcolor{green!65!black}{\downarrow 7.5}}$} \\
    Bridge & 29.1 & 54.5 & 26.0 & 19.3\rlap{$_{\textcolor{green!65!black}{\downarrow 9.8}}$} & 43.0\rlap{$_{\textcolor{green!65!black}{\downarrow 11.5}}$} & 14.8\rlap{$_{\textcolor{green!65!black}{\downarrow 11.2}}$} \\
    Service-area & 65.1 & 92.6 & 76.5 & 60.0\rlap{$_{\textcolor{green!65!black}{\downarrow 5.1}}$} & 88.2\rlap{$_{\textcolor{green!65!black}{\downarrow 4.4}}$} & 69.1\rlap{$_{\textcolor{green!65!black}{\downarrow 7.4}}$} \\
    Toll-station & 58.3 & 82.5 & 61.7 & 56.5\rlap{$_{\textcolor{green!65!black}{\downarrow 1.8}}$} & 82.5 & 59.6\rlap{$_{\textcolor{green!65!black}{\downarrow 2.1}}$} \\
    Harbor & 39.1 & 57.1 & 43.6 & 24.7\rlap{$_{\textcolor{green!65!black}{\downarrow 14.4}}$} & 38.0\rlap{$_{\textcolor{green!65!black}{\downarrow 19.1}}$} & 26.9\rlap{$_{\textcolor{green!65!black}{\downarrow 16.7}}$} \\
    Overpass & 42.7 & 67.0 & 45.4 & 36.7\rlap{$_{\textcolor{green!65!black}{\downarrow 6.0}}$} & 60.6\rlap{$_{\textcolor{green!65!black}{\downarrow 6.4}}$} & 38.2\rlap{$_{\textcolor{green!65!black}{\downarrow 7.2}}$} \\
    Ship & 54.4 & 89.4 & 61.3 & 51.1\rlap{$_{\textcolor{green!65!black}{\downarrow 3.3}}$} & 84.2\rlap{$_{\textcolor{green!65!black}{\downarrow 5.2}}$} & 57.3\rlap{$_{\textcolor{green!65!black}{\downarrow 4.0}}$} \\
    Trainstation & 34.7 & 69.6 & 30.3 & 31.2\rlap{$_{\textcolor{green!65!black}{\downarrow 3.5}}$} & 64.3\rlap{$_{\textcolor{green!65!black}{\downarrow 5.3}}$} & 28.6\rlap{$_{\textcolor{green!65!black}{\downarrow 1.7}}$} \\
    Vehicle & 31.8 & 58.8 & 30.8 & 28.5\rlap{$_{\textcolor{green!65!black}{\downarrow 3.3}}$} & 53.0\rlap{$_{\textcolor{green!65!black}{\downarrow 5.8}}$} & 27.3\rlap{$_{\textcolor{green!65!black}{\downarrow 3.5}}$} \\
    \bottomrule
  \end{tabular}
  }
\end{table}

\subsection{Limitations and Future Work}\label{Limitations}
\begin{figure}
    \centering
    \includegraphics[width=1\linewidth]{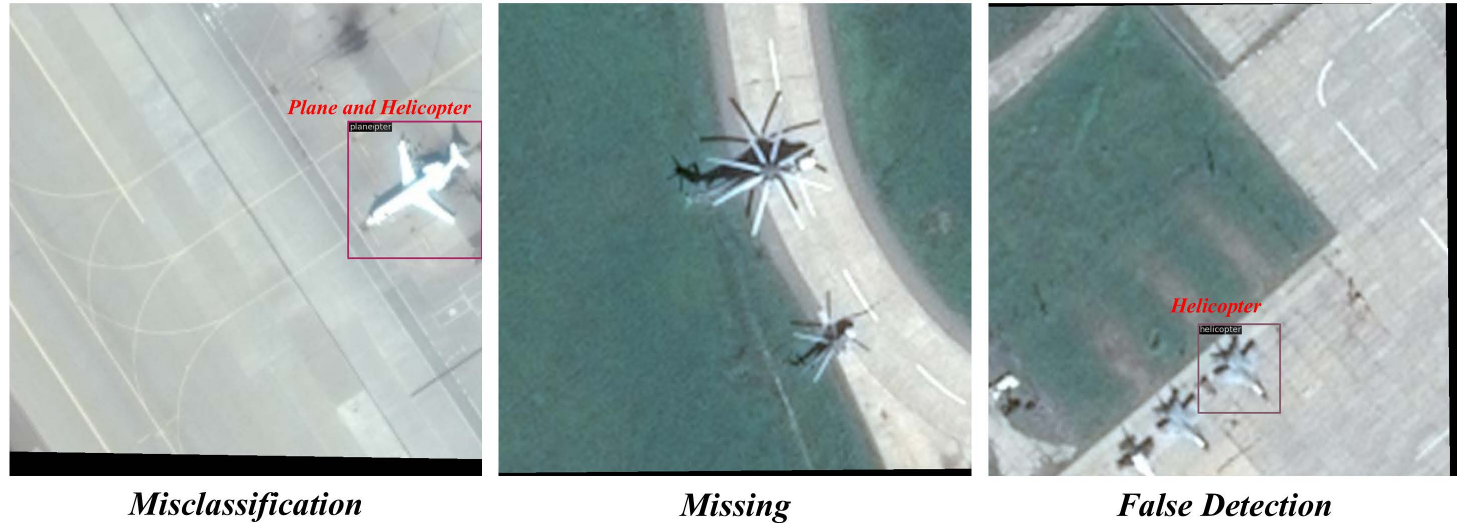}
    \caption{Visualization of typical failure cases for the Helicopter category in the DOTA-IOD dataset. (Left) Misclassification: Planes are erroneously classified as Helicopters due to the similarity of text embeddings between the two categories. (Middle) Missed Detection: True Helicopter instances are missed by the detector. (Right) False Detection: The model incorrectly predicts Helicopters on ambiguous background regions or other objects.}
    \label{limi}
\end{figure}
While STAR-IOD proves effective in mitigating catastrophic forgetting overall, we identify a failure case for the Helicopter category on the DOTA-IOD dataset. Specifically, the $AP$ collapses from a joint-training baseline of 36.7\% to a mere 0.9\% in the incremental setting. As illustrated in Fig.~\ref{limi}, the model struggles to discriminate between Helicopters and Planes. Furthermore, in Task 2 of DOTA-IOD, the $AP_{50}$ for Helicopter reaches only 3.6\%, significantly lagging behind the 64.3\% achieved under joint training. We attribute this performance degradation to the high semantic similarity of text embeddings in the semantic space. Built upon the Grounding DINO architecture, STAR-IOD relies on text prompts for detection. However, the tokens for Plane and Helicopter exhibit high proximity within the text embedding space. Given the non-salient visual features of small-scale helicopters, the model struggles to orthogonalize their representations in the feature space based solely on subtle embedding differences, resulting in ambiguous decision boundaries.

In future work, we plan to incorporate more powerful text encoders to enhance the discriminability and separability of text embeddings in the semantic space, thereby alleviating confusion between semantically similar categories and further improving the robustness and generalization of the model in complex RS-IOD scenarios. 
\section{Conclusion}\label{seccon}
Conventional object detection paradigms are prone to catastrophic forgetting during incremental learning phases, a vulnerability significantly exacerbated by the challenges of remote sensing imagery—specifically, scale variations and background noise interference. To address these challenges, this paper conducts a systematic investigation into RS-IOD. Specifically, we construct two standardized RS-IOD benchmarks, namely DIOR-IOD and DOTA-IOD, which provide a solid foundation for future research in this field. Building upon these benchmarks, we propose a novel framework named STAR-IOD.
STAR-IOD designs a subspace-decoupled topological distillation strategy that aligns the topological structures of knowledge within scale-specific subspaces, alleviating feature distribution shifts caused by scale variations and enabling stable transfer of old knowledge.
Furthermore, we design a plug-and-play pseudo-label generator that adaptively produces high-quality pseudo-labels for old classes via clustering, thereby maintaining foreground consistency throughout the incremental learning stages.
Extensive experimental results on the DOTA-IOD and DIOR-IOD datasets demonstrate that STAR-IOD effectively mitigates catastrophic forgetting, outperforming current SOTA methods by margins of {2.1\%} and {1.7\%} in $mAP$, respectively. These results comprehensively validate the effectiveness and superiority of the proposed method in handling complex remote sensing incremental detection tasks.

\bibliographystyle{model5-names}

\bibliography{main}


\end{document}